\documentclass[letterpaper]{article} % DO NOT CHANGE THIS
\usepackage{aaai24}  % DO NOT CHANGE THIS
\usepackage{times}  % DO NOT CHANGE THIS
\usepackage{helvet}  % DO NOT CHANGE THIS
\usepackage{courier}  % DO NOT CHANGE THIS
\usepackage[hyphens]{url}  % DO NOT CHANGE THIS
\usepackage{graphicx} % DO NOT CHANGE THIS
\usepackage{natbib}  % DO NOT CHANGE THIS AND DO NOT ADD ANY OPTIONS TO IT
\usepackage{caption} % DO NOT CHANGE THIS AND DO NOT ADD ANY OPTIONS TO IT
\usepackage{algorithm}
\usepackage{algorithmic}

\usepackage{amsthm}
\usepackage{amssymb}
\usepackage{color}  %ADD	
\usepackage{latexsym}
\usepackage{multirow}

\newtheorem{definition}{Definition}

\newtheorem{theorem}{Theorem}
\newtheorem{example}{Example}

\newcommand{\set}[1]{\{ #1 \}}

\newcommand{\lang}{\mathcal{L}}
\newcommand{\varSet}{\mathcal{P}}

\newcommand{\belSet}{\psi}
\newcommand{\worldSet}{\mathcal{M}}
\newcommand{\subWorldSet}{\mathcal{N}}
\newcommand{\subWorldSetH}{\mathcal{H}}
\newcommand{\belState}{S}
\newcommand{\belStatus}{S}
\newcommand{\staTObel}[1]{\mathtt{B}({#1})} %\mbox{\textit{B}}({#1})}
\newcommand{\belRe}{\circ}
\newcommand{\belsetUpd}{\diamond} 
\newcommand{\belUpd}{\triangledown}%{\scriptstyle \lozenge}}
\newcommand{\belUpdOper}{\belUpd} %{\scriptstyle \blacklozenge}}%{\blacklozenge}
\newcommand{\staRe}{\circ}
\newcommand{\true}{\top}
\newcommand{\false}{\bot}
\newcommand{\size}[1]{|{#1}|}

\newcommand{\nmodels}{\not \models}

\newcommand{\form}[1]{\mathtt{Form}({#1})} %\mbox{\textit{form}}({#1})}

\newcommand\ie{{\it i.e.}}

\usepackage{newfloat}
\usepackage{listings}
\DeclareCaptionStyle{ruled}{labelfont=normalfont,labelsep=colon,strut=off} % DO NOT CHANGE THIS
\floatstyle{ruled}
\newfloat{listing}{tb}{lst}{}
\floatname{listing}{Listing}
 \nocopyright %-- Your paper will not be published if you use this command
\title{An Investigation of Darwiche and Pearl's Postulates for Iterated Belief Update}

\author {
    Quanlong Guan\textsuperscript{\rm 1},
    Tong Zhu\textsuperscript{\rm 1},
    Liangda Fang\textsuperscript{\rm 1}\thanks{Corresponding author},
    Junming Qiu\textsuperscript{\rm 1},
    Zhao-Rong Lai\textsuperscript{\rm 1},
    Weiqi Luo\textsuperscript{\rm 1}
}
\affiliations {
    \textsuperscript{\rm 1}Jinan University, Guangzhou 510632, China\\
     Ggl@jnu.edu.cn, zhutong62@stu2020.jnu.edu.cn, fangld@jnu.edu.cn, 2040697476jnu@stu2020.jnu.edu.cn, laizhr@jnu.edu.cn, lwg@jnu.edu.cn
}

\usepackage{bibentry}
\begin{document}

\maketitle

\begin{abstract}
	\looseness=-1
	Belief revision and update, two significant types of belief change, both focus on how an agent modify her beliefs in presence of new information.
	The most striking difference between them is that the former studies the change of beliefs in a static world while the latter concentrates on a dynamically-changing world.
	The famous AGM and KM postulates were proposed to capture rational belief revision and update, respectively.
	However, both of them are too permissive to exclude some unreasonable changes in the iteration.
	In response to this weakness, the DP postulates and its extensions for iterated belief revision were presented.
	Furthermore, \citeauthor{Rod2015} integrated these postulates in belief update.
	Unfortunately, his approach does not meet the basic requirement of iterated belief update.
	This paper is intended to solve this problem of \citeauthor{Rod2015}'s approach.
	Firstly, we present a modification of the original KM postulates based on belief states. 
	Subsequently, we migrate several well-known postulates for iterated belief revision to iterated belief update.
	Moreover, we provide the exact semantic characterizations based on partial preorders for each of the proposed postulates.
	Finally, we analyze the compatibility between the above iterated postulates and the KM postulates for belief update.
\end{abstract}

	\vspace*{-2mm}
	\section{Introduction}
	\vspace*{-1mm}
	%\looseness=-1
	Belief revision focuses on how an agent change her beliefs when she encounters new information inconsistent with her initial beliefs.
	The notable AGM postulates, proposed by \citet{AlcGM1985}, have become a standard framework to capture the rational behavior of belief revision.
	\citet{KatM1991a} proposed a characterization of all revision operators that satisfy AGM postulates in terms of total preorders over possible worlds.
	
	\looseness=-1
	Belief update, another significant type of belief change, concentrates on how an agent will modify her beliefs about a dynamically-changing world in view of new information. 
	As in belief revision, \citet{KatM1991b} proposed the KM postulates for regulating rational belief update, which models the process of update as a function of belief sets.
	Furthermore, they offered a semantic characterization based on partial preorders over possible worlds, and clarified the distinctions between update and revision from the model-theoretic perspective. %$ of the preorder-based representation.
	There is only one total preorder for the belief set $\belSet$ in belief revision.
	In contrast, in belief update, a collection of partial preorders is induced by $\belSet$ where each preorder is associated to each possible world satisfying $\belSet$. % is associated with a preorder. % (??)

	\looseness=-1
	Although the AGM postulates have been considered as a basic framework for belief revision, it is shown to be too permissive to exclude some unreasonable revision operators in the iteration \cite{DarP1997}.
	The reason can be attributed to the fact that it is comprised of merely a set of one-step postulates, failing to properly deal with the sequential new information in the process of iterated belief revision.
	To remedy this defect in belief revision, \citeauthor{DarP1997} supplemented the AGM paradigm with four postulates (C1)-(C4) (called DP postulates) and use belief states to denote the belief of an agent instead of belief sets.	
	%	replaced belief sets with belief states.
	Two different belief states may have the same belief sets, but not vice versa.

	Likewise, we argue that the same problem as above exists for the KM postulates in belief update.
	In details, the KM framework is unable to regulate the preferences for subsequent updates during the iterated update process, leading to some counter-intuitive results. % during belief update.
	We use the following examples to briefly illustrate the problem:
	\begin{example} \label{exm:KM_irrational}
		Consider a table with two zones: left and right.
		There is a book, a novel and a toy on any side on the table.
		We denote by $b$ (resp. $n$ / $t$) the proposition ``the book (resp. novel / toy) is on the left zone of the table".
		Initially, everything is on the right zone, and hence the initial belief set $\belSet$ is $\neg b \land \neg n \land \neg t$.
		We first instruct a robot to move at least one of the book and the novel to the left zone, which is described as the new information $\phi = b \lor n$.
		Suppose that updated belief set $\belSet \belsetUpd \phi$ is $(b \leftrightarrow \neg n) \land \neg t$, that is, exactly one of the book and novel is on the left zone, and the toy is on the right zone.
		The robot is then given a fresh instruction to place the novel on the left, which is the new information $\varphi =  n$.
		According to the KM postulates, it is acceptable that the updated belief set $(\belSet \belsetUpd \phi) \belsetUpd \varphi$ becomes $\neg b \land n$.
		However, after updating by $\phi$, we believe that the toy was on the right zone.
		It seems unreasonable to require the agent to give up this belief after putting  the novel on the left zone. \qed
	\end{example}
	
	\looseness=-1
	The two updated beliefs $\belSet \belsetUpd \phi$ and $(\belSet \belsetUpd \phi) \belsetUpd \varphi$ in the above examples are all KM-compatible.
	Since the KM postulates do not constrain the update strategy after iteration, counter-intuitive results will emerge after multiple iterations.
	
	\looseness=-1
	One topic worth investigation is what will happen if we incorporate the iterated revision postulates above to a belief update scenario.
	Following the semantic characterization for every iterated postulate in belief revision, \citet{Rod2015} integrated these postulates in belief update.
	However, \citeauthor{Rod2015}'s approach requires that the operator updates the initial belief set with different preferences over possible worlds no matter how much new information comes.
	For example, an agent has the initial belief $\belSet$ and the preference about the worlds $\leq$.
	She then receives two sequential new pieces of information $\phi$ and $\varphi$.
	In the first phase, the new belief $\belSet \belsetUpd \phi$ together with the new preference $\leq'$ are obtained according to the original belief $\belSet$, the preference $\leq$ and the new information $\phi$.
	However, in \citeauthor{Rod2015}'s approach, the new belief $(\belSet \belsetUpd \phi) \belsetUpd \varphi$ is constructed from the initial belief set $\belSet$, the new preference $\leq'$ and the sentence $\varphi$.
	Clearly, it is more reasonable that the belief $(\belSet \belsetUpd \phi) \belsetUpd \varphi$ should come from the latest belief $\belSet \belsetUpd \phi$ rather than the original one $\belSet$. %every time when receiving the new information.	
	The operator defined by \citet{Rod2015} obviously does not meet the basic requirement of iterated belief update.

	\looseness=-1
	This paper is intended to fix the deficiency of \citeauthor{Rod2015}'s approach.
	The main contributions are the following.
	We first present a modification of the original KM postulates, allowing belief update to be a function of belief states.
	Subsequently, we extend the DP postulates for iterated belief revision to iterated belief update, yielding postulates (C1$\belUpd$)-(C4$\belUpd$). %, (Nat$\belUpd$), (Lex$\belUpd$) and (Ind$\belUpd$). 
	In addition, every postulate requires that the update should result from the latest belief.
	Furthermore, we provide the exact semantic characterizations based on partial preorders for every postulate of iterated belief update.
	Finally, we analyze the compatibility between the above iterated postulates and the KM postulates for belief update.
	We identify an update operator that satisfies (C3$\belUpd$) and (C4$\belUpd$). %, (Lex$\belUpd$) and (Ind$\belUpd$).
	In particular, we show that each of (C1$\belUpd$) and (C2$\belUpd$) is inconsistent with the KM postulates. %and (Nat$\belUpd$) is inconsistent with the KM postulates.

\vspace*{-2mm}
\section{Formal Preliminaries}
\vspace*{-1mm}
	\looseness=-1
	Throughout this paper, we fix a finite set $\varSet$ of \textit{propositional variables}.
	We define $\lang$ to be the \textit{propositional language} built from $\varSet$, the connectives $\neg$, $\land$ and $\lor$, and two Boolean constants $\true$ (truth) and $\false$ (falsity).

	\looseness=-1
	A propositional sentence $\psi$ is \textit{complete}, iff for every sentence $\varphi \in \lang$, $\belSet \models \varphi$ or $\belSet \models \neg \varphi$.
	A propositional sentence $\psi$ is \textit{consistent}, iff there is no sentence $\varphi \in \lang$ s.t. $\belSet \models \varphi$ and $\belSet \models \neg \varphi$.
	A \textit{possible world} $w$ is a complete consistent set of literals over $\varSet$, \ie, for every $p \in \varSet$, either $p \in w$ or $\neg p \in w$.
	We use $\worldSet$ to denote the set of all possible worlds, and $[\phi]$ to denote the set of all possible worlds in which $\phi$ holds.
	For a finite set of worlds $W$, $\form{W}$ denotes the sentence $\bigvee_{w \in W} \left(\bigwedge_{l \in w} l\right)$.
	For ease of representation, we sometimes use $w_1, \cdots, w_n$ to denote the set $\set{w_1, \cdots, w_n}$.
	
	\looseness=-1
	A \textit{(partial) preorder} $\leq$ over $\worldSet$ is a reflexive, transitive binary relation on $\worldSet$. 
	A preorder is \textit{total} if for all $w, w' \in \worldSet$, either $w \leq w'$ or $w' \leq w$. 
	We define $<$ as the strict part of $\leq$, \ie, $w < w'$ iff $w \leq w'$ and $w' \not\le w$. 
	We define $\approx$ as the symmetric part of $\leq$, \ie, $w \approx w'$ iff $w \leq w'$ and $w' \leq w$.

\vspace*{-2mm}
\section{Background}
\vspace*{-1mm}
	% This section briefly introduces the AGM paradigm proposed by \cite{AlcGM1985}, an ideal framework that captures the notion of rational belief revision, and followed by a famous extensions: iterated belief revision \cite{DarP1997,NayPP2003} and relevance-sensitive belief revision \cite{Par1999,PepWCF2015}.
	In this section, we briefly review the AGM postulates for belief revision \cite{AlcGM1985}, the DP postulates for iterated belief revision \cite{DarP1997}, and the KM postulates for belief update \cite{KatM1991b}. % \cite{Bou1996,DarP1997,NayPP2003,BooM2006,JinT2007}, and the KM postulates for belief update \cite{KatM1991b}.
	
	\subsection{The Modified AGM Postulates on Belief States}
%	\vspace*{-1mm}
	\looseness=-1
	The original AGM paradigm models the notion of belief revision as a function that maps a belief set $\belSet$ and a sentence $\phi$ to a new belief set $\belSet \staRe \phi$.
	%As pointed out in \cite{AlcGM1985}, AGM postulates is the framework that all rational revision functions must satisfy.
	In this paper, a belief set is defined as a propositional sentence.
	However, the revision function over belief sets only differentiates what the agent believes and what she does not believe, but does not compare the plausibility degree of different information which the agent does not believe.
	This leads to improper behaviors of revision functions over belief sets on iterated belief revision \cite{DarP1997}.
	To fix this defect, \citeauthor{DarP1997} proposed the notion of belief states (also referred to as epistemic states), redefined revision function on belief states, and reformulated the AGM postulates accordingly.
	\citeauthor{DarP1997} did not provide a standard definition of belief states, and only required that each belief state $\belState$ is associated with a belief set $\staTObel{\belState}$.
	%	For ease of presentation, in the rest of this paper, we identify each belief state $\belState$ as a total preorder $\leq^{\belState}$ over possible worlds, which is one of the classical definitions of belief states\footnote{In this paper, we do not differentiate between ``belief state" and ``total preorder", and use these notions interchangeably.}.		

	We use this abstract representation of the belief state herein.

		The following are the modified AGM postulates which shift from belief sets to belief states and are originated from \cite{DarP1997}.
		\begin{description}%\tightlist
			\item[(R*1)] $\staTObel{\belState \staRe \varphi}$ implies $\varphi$.
			\item[(R*2)] If $\staTObel{\belState} \land \varphi$ is consistent, then $\staTObel{\belState \staRe \varphi} \equiv \staTObel{\belState} \land \varphi$.
			\item[(R*3)] If $\varphi$ is consistent, so is $\staTObel{\belState \staRe \varphi}$.
			\item[(R*4)] If $\belState_1 = \belState_2$ and $\varphi \equiv \phi$, then $\staTObel{\belState_1 \staRe \varphi} \equiv \staTObel{\belState_2 \staRe \phi}$.
			\item[(R*5)] $\staTObel{\belState \staRe \varphi} \land \phi \models \staTObel{\belState \staRe (\varphi \wedge \phi)}$.
			\item[(R*6)] If $\staTObel{\belState \staRe \varphi} \land \phi$ is consistent, then $\staTObel{\belState \staRe (\varphi \wedge \phi)} \models \staTObel{\belState \staRe \varphi} \land \phi$.
		\end{description}

	\citeauthor{DarP1997} provided a representation theorem for the modified AGM postulates based on the notion of faithful total preorders and faith assignments proposed by \citet{KatM1991a}.
	Given a sentence $\varphi$, a total preorder $\leq_{\varphi}$ over $\worldSet$ is \textit{faithful to $\varphi$}, iff (1) $w \approx_{\varphi} w'$ for every two possible worlds $w, w' \in [\varphi]$; and (2) $w <_{\varphi} w'$ for every two possible worlds $w \in [\varphi]$ and $w' \in [\neg \varphi]$.
	A \textit{faithful assignment} is a function that maps each belief state $\belState$ to a total preorder $\leq_{\belState}$ that is faithful to $\staTObel{\belState}$.
	The following is the representation theorem for the modified AGM postulates.
	\begin{theorem}[\cite{DarP1997}] \label{thm:revisionRepresentation} 
		A revision operator $\staRe$ satisfies postulates (R*1)-(R*6) iff there exists a faithful assignment that maps each belief state $S$ to a total preorder $\leq_{\belState}$ s.t. $[\staTObel{\belState \staRe \varphi}] = \min([\varphi], \leq_{\belState})$.
%		\begin{center}\vspace*{-3mm}
%			\item[(F$\staRe$)] $[\belState \staRe \varphi] = \min([\varphi], \leq_{\belState})$.
%		\end{center}%\vspace*{-1mm}
	\end{theorem}

\subsection{Iterated Belief Revision} \label{sec:iterated_revision_postulates}

	\looseness=-1
	In the AGM paradigm, there is no guidance on how to obtain the relationship between the initial revision strategy and the subsequent one.
	To solve this problem, \citeauthor{DarP1997} proposed DP postulates that describe rational iterated belief revision.
		\begin{description}
			\item[(C1)] If $\varphi \models \phi$, then $\staTObel{(\belState \staRe \phi) \staRe \varphi} \equiv \staTObel{\belState \staRe \varphi}$.
			\item[(C2)] If $\varphi \models \neg\phi$, then $\staTObel{(\belState \staRe \phi) \staRe \varphi} \equiv \staTObel{\belState \staRe \varphi}$.
			\item[(C3)] If $\staTObel{\belState \staRe \varphi} \models \phi$, then $\staTObel{(\belState \staRe \phi) \staRe \varphi} \models \phi$.
			\item[(C4)] If $\staTObel{\belState \staRe \varphi} \not \models \neg \phi$, then $\staTObel{(\belState \staRe \phi) \staRe \varphi} \not \models \neg \phi$.
		\end{description}

	\looseness=-1
	\citeauthor{DarP1997} proved that postulates (C1)-(C4) correspond to the following semantics constraints (CR1)-(CR4) on possible worlds, respectively. %, associating the preorder ⪯S of the initial belief state S with the preorder ⪯S◦ϕ of	the belief state that results from the revision of S by ϕ.
	
	%The semantic counterparts of the four DP postulates and the representation theorem are given in the following.
	\begin{description}
		\item[(CR1)] If $w_1, w_2 \in [\phi]$, then $w_1 \leq_{\belState} w_2$ iff $w_1 \leq_{\belState \staRe \phi} w_2$.
		\item[(CR2)] If $w_1, w_2 \in [\neg \phi]$, then $w_1 \leq_{\belState} w_2$ iff $w_1 \leq_{\belState \staRe \phi} w_2$.
		\item[(CR3)] If $w_1 \in [\phi]$ and $w_2 \in [\neg \phi]$, then $w_1  <_{\belState} w_2$ only if $w_1 <_{\belState \staRe \phi} w_2$.
		\item[(CR4)] If $w_1 \in [\phi]$ and $w_2 \in [\neg \phi]$, then $w_1 \leq_{\belState} w_2$ only if $w_1 \leq_{\belState \staRe \phi} w_2$.
	\end{description}
	
	\begin{theorem}[\cite{DarP1997}] %\label{thm:iterRepr}
		Let $\staRe$ be a revision operator satisfying (R*1)-(R*6).
		Then, $\staRe$ satisfies (Ci) iff the operator and its corresponding faithful assignment satisfies (CRi) for $1 \leq i \leq 4$.
		
%		the following hold
%		\begin{itemize}
%			\item $\staRe$ satisfies (C1) iff the operator and its corresponding faithful assignment satisfies (CR1).
%			\item $\staRe$ satisfies (C2) iff the operator and its corresponding faithful assignment satisfies (CR2).
%			\item $\staRe$ satisfies (C3) iff the operator and its corresponding faithful assignment satisfies (CR3).
%			\item $\staRe$ satisfies (C4) iff the operator and its corresponding faithful assignment satisfies (CR4).
%		\end{itemize} 
	\end{theorem}

\subsection{The KM Postulates for Belief Update}
%\vspace*{-1mm}

	\looseness=-1
	In the belief update literature, an agent is intended to modify her beliefs about a dynamically-changing environment in view of new information.
	\citet{KatM1991b} clarified the distinction between belief revision and update.
	Furthermore, following the AGM paradigm, they presented the KM postulates to characterize a family of rational belief update functions, which map a belief set $\belSet$ and a sentence $\varphi$ to a new belief set $\belSet \belsetUpd \varphi$.
	
		\begin{description}
			\item[(U1)] $\belSet \belsetUpd \varphi$ implies $\varphi$.
			\item[(U2)] If $\belSet$ implies $\varphi$, then $\belSet \belsetUpd \varphi \equiv \belSet$.
			\item[(U3)] If both $\belSet$ and $\varphi$ are consistent, so is $\belSet \belsetUpd \varphi$.
			\item[(U4)] If $\varphi \equiv \phi$, then $\belSet \belsetUpd \varphi \equiv \belSet \belsetUpd \phi$.
			\item[(U5)] $(\belSet \belsetUpd \varphi) \land \phi$ implies $\belSet \belsetUpd (\varphi \wedge \phi)$.
			\item[(U6)] If $\belSet \belsetUpd \varphi$ implies $\phi$ and $\belSet \belsetUpd \phi$ implies $\varphi$, then $\belSet \belsetUpd \varphi \!\equiv\! \belSet \belsetUpd \phi$.
			\item[(U7)] If $\belSet$ is complete, then $(\!\belSet \belsetUpd \varphi) \!\land\! (\belSet \belsetUpd \phi)$ implies $\belSet \belsetUpd (\varphi \lor \phi)$.
			\item[(U8)] $(\belSet_1 \lor \belSet_2) \belsetUpd \varphi \equiv (\belSet_1 \belsetUpd \varphi) \lor (\belSet_2 \belsetUpd \varphi)$.
			
		\end{description}

	\looseness=-1
	To describe the process of belief update, \citeauthor{KatM1991b} proposed the notion of \textit{faithful preorder} associated with possible worlds.
	Formally, given a possible world $w$, a preorder $\leq_{w}$ over $\worldSet$ is \textit{faithful to $w$}, iff for every possible world $w'$, $w' \neq w$ only if $w <_{w} w'$.
	A \textit{faithful pointwise assignment} is a function that maps each possible world $w$ to a partial preorder that is faithful to $w$.
	The following theorem shows the semantic characterization of KM postulates. % based on faithful pointwise assignments.
	\begin{theorem}[\cite{KatM1991b}] \label{thm:KM_semantic} 
		An update operator $\belsetUpd$ satisfies postulates (U1)-(U8) iff there exists a faithful pointwise assignment that maps each possible world $w$ to a partial preorder $\leq_w$ s.t. $[\belSet \belsetUpd \varphi] = \bigcup\limits_{w \in [\belSet]} \min([\varphi],\! \leq_{w})$.
%		\begin{center}\vspace*{-3mm}
%			\item[(F$\belsetUpd$)] $[\belSet \belsetUpd \varphi] = \bigcup\limits_{w \in [\belSet]} \min([\varphi], \leq_{w})$.
%		\end{center}%\vspace*{-1mm}
	\end{theorem}

\section{The Modified KM Postulates on Belief States}

	\looseness=-1
	Similarly to belief revision, in order to regulate the update strategy in the iteration, it is necessary to adopt the notion of belief state rather than belief set as the belief of an agent.

	\looseness=-1
	In the following, we reformulate the KM postulates for belief update in the context of belief state.
	We use $\belStatus \belUpd \varphi$ to denote the resultant belief state via updating the previous one $\belStatus$ by the sentence $\varphi$.	
	
		\begin{description}
			%		\item[(U0')] $\belStatus \belUpd \varphi$ is a belief state. 
			\item[(U1$\belUpd$)] $\staTObel{\belStatus \belUpd \varphi}$ implies $\varphi$.
			\item[(U2$\belUpd$)] If $\staTObel{\belStatus}$ implies $\varphi$, then $\staTObel{\belStatus \belUpd \varphi} \equiv \staTObel{\belStatus}$.
			\item[(U3$\belUpd$)] If both $\staTObel{\belStatus}$ and $\varphi$ are consistent, so is $\staTObel{\belStatus \belUpd \varphi}$.
			\item[(U4$\belUpd$)] If $\varphi \equiv \phi$, then $\staTObel{\belStatus \belUpd \varphi} \equiv  \staTObel{\belStatus \belUpd \phi}$.
			\item[(U5$\belUpd$)] $\staTObel{\belStatus \belUpd \varphi} \land \phi$ implies $\staTObel{\belStatus \belUpd (\varphi \wedge \phi)}$.	
			\item[(U6$\belUpd$)] If $\staTObel{\belStatus \belUpd \varphi}$ implies $\phi$ and $\staTObel{\belStatus \belUpd \phi}$ implies $\varphi$, then $\staTObel{\belStatus \belUpd \varphi} \equiv \staTObel{\belStatus \belUpd \phi}$.
			\item[(U7$\belUpd$)] If $\staTObel{\belStatus}$ is complete, then $\staTObel{\belStatus \belUpd \varphi} \land \staTObel{\belStatus \belUpd \phi}$ implies $\staTObel{\belStatus \belUpd (\varphi \vee \phi)}$.
			\item[(U8$\belUpd$)] $\staTObel{\belStatus \belUpd \varphi} \equiv \bigvee \limits_{w_i \in [\staTObel{\belStatus}]} \staTObel{\belStatus_i \belUpd \varphi}$ for some set of belief states $\set{S_1, \cdots, S_n}$ s.t. $\staTObel{\belStatus_i} \equiv \form{w_i}$.
		\end{description}

		\looseness=-1
		Postulates (U1$\belUpd$)-(U7$\belUpd$) are slightly different from the original KM postulates (U1)-(U7). % except that we use belief state instead.
		Postulates (U8) and (U8$\belUpd$) aim both to achieve the distributive law of the update operator over disjunction although they look quite different at first glance.
		This is because the disjunction connective cannot directly apply to belief state. %However, 
		%We remark that the postulate (U8) and (U8$\belUpd$) are both to achieve the distributive law over the update operator over disjunction, though it seems that the latter is more complicate than the former.
		We illustrate the shift from (U8) to (U8$\belUpd$) in the following.
		Since $\belSet \equiv \bigvee\limits_{w \in [\belSet]} \form{w}$, we can obtain postulate (U8') equivalent to (U8) via iteratively applying (U8).
		\begin{description}
			\item[(U8')] $\belSet \belsetUpd \varphi \equiv \bigvee \limits_{w \in [\belSet]} \left(\form{w} \belsetUpd \varphi \right)$.
		\end{description}
		According to postulate (U8'), the update of any belief set $\belSet$ reduces to the update of $[\belSet]$-possible worlds.
		%To rephrase (U8') in terms of belief states, we use 
		The notations $\staTObel{\belState}$ and $\staTObel{\belStatus \belUpd \varphi}$ are the associated belief sets of the initial belief state $\belState$ and the updated one $\belStatus \belUpd \varphi$, respectively.
		Moreover, the update operator $\belUpd$ takes a belief state and a sentence as input.
		In order to describe the update of each possible world $w_i$ of $\staTObel{\belState}$, we choose a belief state $S_i$ such that the model of its associated belief set $\staTObel{S_i}$ is exactly $w_i$. % so as to reflect the update of $w_i$.
		Hence, postulate (U8$\belUpd$) coincides with (U8) and (U8').

	\looseness=-1
	We hereafter provide the definition of faithful collective assignments over belief states.
	For ease of presentation, we use $\leq^{\belStatus}$ for a collection $\set{\leq^{\belStatus}_{w}}_{w \in [\staTObel{S}]}$ of partial preorders. % over $\worldSet$.	
	
	\begin{definition} \rm  \label{def:faithPointwiseAssgin}
		A \textit{faithful collective assignment} is a function that maps each belief state $\belStatus$ to a collection $\leq^{\belStatus}$ of partial preorders s.t. for each belief state $\belStatus$ and each possible world $w \in [\staTObel{\belStatus}]$, we have that 
		\begin{itemize}
			\item each preorder $\leq^{\belStatus}_{w}$ is faithful to $w$; and
			
			%			for every $w' \in \worldSet$, $w' \neq w$ implies $w <^{\belStatus}_{w} w'$; 
			\item there is a belief state $\belStatus'$ s.t. $\staTObel{\belStatus'} \equiv \form{w}$ and $\leq^{\belStatus}_{w} = \leq^{\belStatus'}_{w}$.
		\end{itemize}
	\end{definition}

	\looseness=-1
	We point out the difference among faithful assignments for (iterated) belief revision, faithful pointwise assignments for belief update and faithful collective assignments for iterated belief update.
	A faithful assignment assigns each belief state to a single total preorder and a faithful pointwise assignment maps each possible world to a single partial preorder.
	In contrast, a faithful collective assignment associates each belief state with a collection of partial preorders.

	\looseness=-1
	We hereafter show that Theorem \ref{thm:KM_semantic} can be extended to the modified KM postulates and faithful collective assignments. % faithful. % to belief states.

	\begin{theorem} \label{thm:ModifiedKM_semantic}
		% An operator $\belUpd$ is an update operator iff there exists a faithful collective assignment that maps each belief state $\belStatus$ and possible world $w \in \worldSet$ to a total preorder $\leq^{\belStatus}_{w}$ s.t. 
		An update operator $\belUpd$ satisfies postulates (U1$\belUpd$)-(U8$\belUpd$) iff there exists a faithful collective assignment that maps each belief state $\belStatus$ to a collection $\leq^{\belStatus}$ of partial preorders s.t. $[\staTObel{\belStatus \belUpd \varphi}] = \bigcup\limits_{w \in [{\staTObel{\belStatus}}]} \min([\varphi], \leq^{\belStatus}_{w})$.
%		\begin{center}\vspace*{-3mm}
%			\item[(F$\belUpd$)] $[\staTObel{\belStatus \belUpd \varphi}] = \bigcup\limits_{w \in [{\staTObel{\belStatus}}]} \min([\varphi], \leq^{\belStatus}_{w})$.
%		\end{center}%\vspace*{-1mm}
	\end{theorem}

%	\vspace*{-2mm}
	\section{Iterated Belief Update on Belief States}
%	\vspace*{-1mm}
	\looseness=-1
	We are ready to investigate the iteration of belief update with the help of belief states.
	In this section, we first modify the iterated revision postulates so as to suit the update operator, and then examine the rationale of the proposed iterated update postulates with concrete examples. %real-life scenarios.
	%In addition, we  
	Moreover, we provide the model-theoretic characterization of each of the DP postulates in the context of belief update and prove the representation theorem. % for iterated belief update.
	Finally, we make a brief comparison between our approach and \citeauthor{Rod2015}'s. % one \cite{Rod2015}. %, illustrating how we fix the deficiency in \citeauthor{Rod2015}'s approach.

	%\vspace*{-2mm}
	\subsection{Iterated Update Postulates and Examples}
%	\vspace*{-1mm}
	\looseness=-1
	The iterated postulates can be easily transferred to belief update via replacing the revision operator $\belRe$ by the update one $\belUpd$.
	To distinguish the iterated postulates for belief update and revision, we add a symbol $\belUpd$ behind the name of the original iterated revision postulate for the corresponding one in iterated belief update.
	For example, (C1$\belUpd$) is ``if $\varphi \models \phi$, then $\staTObel{(\belStatus \belUpd \phi) \belUpd \varphi} \equiv \staTObel{\belStatus \belUpd \varphi}$" while (C1) is ``if $\varphi \models \phi$, then $\staTObel{(\belStatus \belRe \phi) \belRe \varphi} \equiv \staTObel{\belStatus \belRe \varphi}$".
	
	\begin{example}[Postulate (C1$\belUpd$)] \label{exm:C1}
		\looseness=-1
		Let us continue Example \ref{exm:KM_irrational}.
		Let $\varSet = \set{b, n, t}$ and $\worldSet = \set{w_0, \dots, w_7}$.
		The definition of each possible world is shown in Table \ref{tab:KM_operator_violating_C1}, in which an occurrence of $\true$ (resp. $\false$) indicates that the positive (resp. negative) literal of the corresponding proposition is in the possible world.				
		For example, all of the three cells of the rows ``book", ``novel" and ``toy" and the column ``$w_0$" are $\false$, meaning that they are all on the right zone of the table in the possible world $w_0$.
		The other possible worlds are defined similarly.
		We provide a KM-compatible update operator in Example \ref{exm:KM_irrational} as follows.
		The initial belief state $\belStatus$ is assigned to the set $\leq^{\belStatus}$ of partial preorders with one element $\leq^{\belStatus}_{w_0}$.
		\begin{itemize}
			\item $w_0 <^{\belStatus}_{w_0} w_2, w_4 <^{\belStatus}_{w_0} w_1, w_3, w_5, w_6, w_7$.
		\end{itemize}
		
		\looseness=-1
		Then, the updated belief state $\belStatus \belUpd \phi$ are assigned the collection $\leq^{\belStatus \belUpd \phi}$ of two preorders listed below.
		\begin{itemize}
			\item $w_2 <^{\belStatus \belUpd \phi}_{w_2} w_4 <^{\belStatus \belUpd \phi}_{w_2} w_0, w_1, w_3, w_5, w_6, w_7 $.
			\item $w_4 <^{\belStatus \belUpd \phi}_{w_4} w_3 <^{\belStatus \belUpd \phi}_{w_4}  w_0, w_1, w_2, w_5, w_6, w_7$.
		\end{itemize}
		
		\looseness=-1
		The two worlds $w_2$ and $w_3$ are the minimal element satisfying $\varphi$ w.r.t. $\leq^{\belStatus \belUpd \phi}_{w_2}$ and $\leq^{\belStatus \belUpd \phi}_{w_4}$, respectively.		
		According to the semantics of the update operator (cf. Theorem \ref{thm:ModifiedKM_semantic}), we have that $\staTObel{(\belStatus \belUpd \phi) \belUpd \varphi} \equiv \form{w_2, w_3} \equiv \neg b \land n$, discarding unjustifiably the belief that the toy is on the right zone. %, due to $w_3 <^{\belStatus \belUpd \phi}_{w_4} w_2$.
		
		\looseness=-1
		In contrast, from (C1$\belUpd$), we deduce that $\staTObel{(\belStatus \belUpd \phi) \belUpd \varphi} \equiv \staTObel{\belStatus \belUpd \varphi} \!\!\equiv\!\! \neg b \land n \land \neg t$, preserving the belief about the toy. \qed
	\end{example}

	\begin{table}[t] 
		\centering
		%\resizebox{\columnwidth}{!}{
			\footnotesize
			\begin{tabular}{ll|l|l|l|l|l|l|l|l|}
				\hline
				\multicolumn{2}{|l|}{possible worlds}
				& $w_0$    & $w_1$    & $w_2$    & $w_3$    & $w_4$    & $w_5$    & $w_6$    & $w_7$   \\ \hline \hline
				\multicolumn{2}{|l|}{\textbf{b}ook}                                                                                
				& $\false$ & $\false$ & $\false$ & $\false$ & $\true$  & $\true$  & $\true$  & $\true$ \\ \hline
				\multicolumn{2}{|l|}{\textbf{n}ovel}                                                                            
				& $\false$ & $\false$ & $\true$  & $\true$  & $\false$ & $\false$ & $\true$  & $\true$ \\ \hline
				\multicolumn{2}{|l|}{\textbf{t}oy}                                                                                 
				& $\false$ & $\true$  & $\false$ & $\true$  & $\false$ & $\true$  & $\false$ & $\true$ \\ \hline
				
			\end{tabular}
		%}
		\vspace*{-3mm}
		\caption{The definition of possible worlds in Example \ref{exm:C1}.}
		\vspace*{-3mm}
		\label{tab:KM_operator_violating_C1}
	\end{table}
	
	\begin{example}[Postulates (C2$\belUpd$)]  \label{exm:C2}
		\looseness=-1
		We use the previous example to justify postulate (C2$\belUpd$).
%		The set of possible worlds $\worldSet$ is the same as in the previous example.
		Let us consider the following KM-compatible update operator.
		Initially, the belief set $\staTObel{\belStatus} \equiv \neg b \land \neg n \land \neg t \equiv \form{w_0}$.
		The associated partial preorder $\leq^{\belStatus}_{w_0}$ of the belief state $\belStatus$ is:
		\begin{itemize}
			\item $w_0 <^{\belStatus}_{w_0} w_2, w_4 <^{\belStatus}_{w_0} w_1, w_3, w_5, w_6, w_7$.
		\end{itemize}
		
		\looseness=-1
		Now, a robot is instructed to put the book and the novel on the opposite zone and put the toy on the right zone, described as $\phi = (b \leftrightarrow \neg n) \land \neg t$. 
		From the semantics of the update operator, the updated belief set $\staTObel{\belStatus \belUpd \phi}$ is $(b \leftrightarrow \neg n) \land \neg t \equiv \form{w_2, w_4}$, that is, only one of the book and novel is on the left zone, and the toy must be on the right zone.
		Then, the updated belief state $\belStatus \belUpd \phi$ is assigned the collection $\leq^{\belStatus \belUpd \phi}$ of two partial preorders listed below.
		\begin{itemize}
			\item $w_2 <^{\belStatus \belUpd \phi}_{w_2} w_6,w_7 <^{\belStatus \belUpd \phi}_{w_2} w_0, w_1, w_3, w_4, w_5 $.
			\item $w_4 <^{\belStatus \belUpd \phi}_{w_4} w_0, w_1 <^{\belStatus \belUpd \phi}_{w_4}  w_2, w_3, w_5, w_6, w_7$.
		\end{itemize}
		%			}
		
		\looseness=-1
		Subsequently, the robot is informed to put the novel and the book on the same zone, which is described as new information $\varphi = b \leftrightarrow n$.
		Clearly, the minimal elements satisfying $\varphi$ w.r.t. $\leq^{\belStatus \belUpd \phi}_{w_2}$ (resp. $\leq^{\belStatus \belUpd \phi}_{w_4}$) are possible worlds $w_6$ and $w_7$ (resp. $w_0$ and $w_1$).
		By the semantics of the update operator, the updated belief set $\staTObel{(\belStatus \belUpd \phi) \belUpd \varphi}$ becomes $b \leftrightarrow n \equiv \form{w_0, w_1, w_6, w_7}$.			
		However, as the agent has believed, after the update by $\phi$, that the toy was on the right zone, it seems irrational to require the agent to give up this belief.
		
		\looseness=-1
		As $\varphi \models \neg \phi$, according to postulate (C2$\belUpd$), we can deduce that $\staTObel{(\belStatus \belUpd \phi) \belUpd \varphi} \equiv \staTObel{\belStatus \belUpd \varphi} \equiv \neg b \land \neg n \land \neg t$.
		In this way, it preserves the belief about the toy.
		\qed
	\end{example}

		\begin{example}[Postulate (C3$\belUpd$)] \label{exm:C3}
		\looseness=-1
		Consider the scenario where an alarm is in a warehouse.
		When the alarm sounds, it means a fire breaks out.
		The proposition $a$ and $f$ denotes that ``the alarm sounds" and ``the warehouse catches fire", respectively.
		%\looseness=-1
		Let $\varSet = \set{a, f}$ and $\worldSet = \set{w_0, \dots, w_3}$.
		Each possible world is defined in Table \ref{tab:KM_operator_violating_C3}.
		Initially, the belief set $\staTObel{\belStatus}$ is $\neg a \land \neg f \equiv \form{w_0}$, meaning that neither the alarm sounds nor the warehouse catches fire.
		The belief state $\belStatus$ is associated with the following preorder $\leq^{\belStatus}_{w_0}$. 
		\begin{itemize}
			\item $w_0 <^{\belStatus}_{w_0} w_1 <^{\belStatus}_{w_0} w_3 <^{\belStatus}_{w_0} w_2$.
		\end{itemize}
		
		\looseness=-1
		At some point, some workers smoke, causing a fire in the warehouse, which is described as the new information $\phi = f$.
		%a robot reports that there is a fire in the warehouse, which can be described as the new information $\phi = f$.
		The updated belief set $\staTObel{\belStatus \belUpd \phi}$ is $\neg a \land f \equiv \form{w_1}$, that is, the alarm does not sound but the fire is in the warehouse.
		Then, the updated belief state $\belStatus \belUpd \phi$ is assigned to the set $\leq^{\belStatus \belUpd \phi}$ of preorders with one element $\leq^{\belStatus \belUpd \phi}_{w_1}$.
		\begin{itemize}
			\item $w_1 <^{\belStatus \belUpd \phi}_{w_1} w_0, w_2, w_3$.
		\end{itemize}
		
		\looseness=-1
		Subsequently, the alarm sounds, which is described as new information $\varphi = a$.
		Obviously, the minimal elements satisfying $\varphi$ are possible worlds $w_2$ and $w_3$.
		It follows from the semantics of the update operator that the updated belief set $\staTObel{(\belStatus \belUpd \phi) \belUpd \varphi} \equiv a \equiv \form{w_2, w_3}$.
		However, after updating by $\phi$, we believe that there is a fire in the warehouse.
		It seems unreasonable to give up this belief now when the alarm is also sounding.
		
		\looseness=-1
		As mentioned before, the alarm sounds only when there is a fire, hence $\staTObel{\belStatus \belUpd a} \models f$ holds.
		We infer from postulate (C3$\belUpd$) that $\staTObel{(\belStatus \belUpd \phi) \belUpd \varphi} \models \phi$, allowing to preserve the belief $\phi = f$ in the belief set $\staTObel{(\belStatus \belUpd \phi) \belUpd \varphi}$. 
		\qed
		\end{example}

	\begin{table}[t] 
		\centering
		%\resizebox{\columnwidth}{!}{
			\footnotesize
			\begin{tabular}{ll|l|l|l|l|}
				\hline
				\multicolumn{2}{|l|}{possible worlds}                                                                      
				& $w_0$    & $w_1$    & $w_2$    & $w_3$       \\ \hline \hline
				\multicolumn{2}{|l|}{\textbf{a}larm}                                                                           
				& $\false$ & $\false$ & $\true$ & $\true$  \\ \hline
				\multicolumn{2}{|l|}{\textbf{f}ire}                                                                            
				& $\false$ & $\true$ & $\false$  & $\true$  \\ \hline		
			\end{tabular}
			%}
		\vspace*{-3mm}
		\caption{
			%				A KM-compatible operator violating postulate (C3$\belUpd$) % in Example \ref{exm:C3}.
			The definition of possible worlds in Example \ref{exm:C3}.
		}
		\vspace*{-3mm}
		\label{tab:KM_operator_violating_C3}
	\end{table}
	
	\begin{example}[Postulate (C4$\belUpd$)] \label{exm:C4}
		\looseness=-1
		Consider a scenario similar to Example \ref{exm:C1}.
		The initial belief set $\staTObel{\belStatus}$ is $\neg n \land \neg t \equiv \form{w_0, w_4}$.
		The belief state $\belStatus$ is associated with the set $\leq^{\belStatus}$ of partial preorders that contains two elements $\leq^{\belStatus}_{w_0}$ and $\leq^{\belStatus}_{w_4}$. 
		\begin{itemize}
			\item $w_0 <^{\belStatus}_{w_0} w_4 <^{\belStatus}_{w_0} w_1, w_2, w_3, w_5, w_6, w_7$.
		\end{itemize}
		\begin{itemize}
			\item $w_4 <^{\belStatus}_{w_4} w_0 <^{\belStatus}_{w_4} w_1, w_2, w_3, w_5, w_6, w_7$.
		\end{itemize}
		
		\looseness=-1
		A robot first moves the book to the left zone and hence the new information $\phi = b$.
		Then, it is believed that the book is on the left zone, that is, the updated belief set $\staTObel{\belStatus \belUpd \phi} \equiv b \land \neg n \land \neg t \equiv \form{w_4}$.
		The associated partial preorder $\leq^{\belStatus \belUpd \phi}_{w_4}$ of the belief state $\belStatus \belUpd \phi$ is:
		\begin{itemize}
			\item $w_4 <^{\belStatus \belUpd \phi}_{w_4} w_0, w_2 <^{\belStatus \belUpd \phi}_{w_4} w_1, w_3, w_5, w_6, w_7$.
		\end{itemize}
		
		\looseness=-1
		The robot is subsequently instructed to put the novel on the left zone, which is described as the new information $\varphi = n$.
		Accordingly, we get that the updated belief set $\staTObel{(\belStatus \belUpd \phi) \belUpd \varphi}$ becomes $\neg b \land n \land \neg t \equiv \form{w_2}$.
		However, because $\staTObel{\belStatus \belUpd \phi} \equiv b \land \neg n \land \neg t$, we have no reason to believe that the book is on the right zone. 
		
		\looseness=-1
		Since $\staTObel{\belStatus \belUpd \phi} \not\models \neg\varphi$, postulate (C4$\belUpd$) can be used to rule out the above irrational behaviour.
		Postulate (C4$\belUpd$) requires $\staTObel{(\belState \belUpd \phi) \belUpd \varphi} \nmodels \neg \phi$, forbidding the unreasonable belief $\neg \phi \!=\! \neg b$ to be a consequence of the new belief set.
		\qed
	\end{example}

	\subsection{Representation Theorem}
	\looseness=-1
	We hereafter present the model-theoretic characterization of each of the DP iterated postulates.
	\begin{description}	
		\item[(CR1$\belUpd$)] For every $\subWorldSet \subseteq [\phi]$, the following hold
		\begin{description}
			\item[Forth] for every $w \in [\staTObel{\belStatus}]$ and $w'' \in \min(\subWorldSet,  \leq^\belStatus_w)$, there is $w' \in [\staTObel{\belStatus \belUpd \phi}]$ s.t. $w'' \in \min(\subWorldSet, \leq^{\belStatus \belUpd \phi}_{w'})$.
			\item[Back] \!for every $w' \!\in\! [\staTObel{\belStatus \belUpd \phi}]$ and $w'' \!\in\! \min(\subWorldSet\!,\! \leq^{\belStatus \belUpd \phi}_{w'})$, there is $w \in [\staTObel{\belStatus}]$ s.t. $w'' \in \min(\subWorldSet, \leq^{\belStatus}_{w})$.
		\end{description}
		
		\item[(CR2$\belUpd$)] For every $\subWorldSet \subseteq [\neg \phi]$, the following hold
		\begin{description}
			\item[Forth] for every $w \in [\staTObel{\belStatus}]$ and $w'' \in \min(\subWorldSet,  \leq^\belStatus_w)$, there is $w' \in [\staTObel{\belStatus \belUpd \phi}]$ s.t. $w'' \in \min(\subWorldSet, \leq^{\belStatus \belUpd \phi}_{w'})$.
			\item[Back] \!for every $w' \!\in\! [\staTObel{\belStatus \belUpd \phi}]$ and $w'' \!\in\! \min(\subWorldSet\!,\!  \leq^{\belStatus \belUpd \phi}_{w'})$, there is $w \in [\staTObel{\belStatus}]$ s.t. $w'' \in \min(\subWorldSet, \leq^{\belStatus}_{w})$.
		\end{description}
		
		\item[(CR3$\belUpd$)] For every $\subWorldSet \subseteq \worldSet$ s.t. $\min(\subWorldSet, \leq^\belStatus_w) \subseteq [\phi]$ holds for every $w \in [\staTObel{\belStatus}]$, we have $\min(\subWorldSet,  \leq^{\belStatus \belUpd \phi}_{w'}) \subseteq [\phi]$ for every $w' \in [\staTObel{\belStatus \belUpd \phi}]$.
		
		\item[(CR4$\belUpd$)] For every $\subWorldSet \subseteq \worldSet$ s.t. $\min(\subWorldSet, \leq^\belStatus_w) \cap [\phi] \neq \emptyset$ holds for some $w \in [\staTObel{\belStatus}]$, we have there is $w' \in [\staTObel{\belStatus \belUpd \phi}]$ s.t. $\min(\subWorldSet, \leq^{\belStatus \belUpd \phi}_{w'}) \cap [\phi] \neq \emptyset$.

	\end{description}

	\looseness=-1
	The semantics (CR1$\belUpd$) requires that for every subset $\subWorldSet$ of possible worlds of $\phi$, the minimal elements of $\subWorldSet$ w.r.t. the collection $\leq^{\belStatus \belUpd \phi}$ of preorders associated with the updated belief state $\belStatus \belUpd \phi$ should be in accordance with that of the original one $\belStatus$.
	(CR2$\belUpd$) acts similarly to (CR1$\belUpd$) except that it focuses on the subset of possible worlds of $\neg \phi$.
	The semantics (CR3$\belUpd$) (resp. (CR4$\belUpd$)) states that for every set $\subWorldSet$ of possible worlds, if all of the minimal elements of $\subWorldSet$ w.r.t. the collection $\leq^\belStatus$ of preorders satisfy (resp. falsify) $\phi$, then such property should be retained after updating by the new information $\phi$.

	\looseness=-1
	%	As can be seen above, the semantics of the iterated update postulates are slightlly different from those in belief revision.
	Although each iterated update postulate is identical to the corresponding iterated revision one, the semantics for iterated postulates in belief update is distinct from that in belief revision.
	In contrast to belief revision, which is based on a single preorder, an update strategy is defined as a collection of preorders.
	We formalize the iterated update strategy in terms of the minimal elements of preorders.
	When both of the initial and updated collection of preorders contains a single preorder, we note that the semantics for each iterated update postulate matches with that for the corresponding iterated revision one.

	\looseness=-1
	The theorem below provides the correspondence between each of the above postulates and its corresponding semantics, based on faithful collective assignments.
	
	\begin{theorem} \label{thm:iterRepr}
		Let $\belUpd$ be an update operator satisfying (U1$\belUpd$)-(U8$\belUpd$).
		% Let $f$ be a faithful collective assignment over belief states, corresponding to $\belUpd$ by means of (F$\belUpd$).
		Then, $\belUpd$ satisfies (Ci$\belUpd$) iff the operator and its corresponding faithful collective assignment satisfies (CRi$\belUpd$) for $1 \leq i \leq 4$.
	\end{theorem}

	\subsection{Comparison to Rodrigues's Approach}
	\looseness=-1
	 %(Some terminologies ??)
	\citet{Rod2015} migrated the iterated revision postulates to the update scenarios. %, (??) by following their semantic characterization in revision.
	In his approach, the update operators act on belief sets rather than belief states during the iteration.
	In addition to the standard update operator $\belsetUpd$, he defined a different update operator, denoted by $\belsetUpd_{\phi}$ for each sentence $\phi$.
	The update operator $\belsetUpd_{\phi}$ takes a belief set $\belSet$ and a sentence $\varphi$ as input where $\belSet$ denotes the initial belief set and $\varphi$ denotes the latest information and reflects the behavior of iterated update.
	%	The sequence $s$ is $[\varphi_1, \varphi_2, \cdots, \varphi_n]$ reflecting the iteration of update.
	The notation $\belSet \belsetUpd_{\phi} \varphi$ means that the initial belief set will be updated by $\phi$ and followed by $\varphi$ while $\belSet \belsetUpd \varphi$ denotes the new belief set via updating $\belSet$ by $\varphi$.
	
	\looseness=-1
	The following is the variant of postulate (C1) for iterated belief update and its semantic characterization. %\footnote{In the thesis \cite{Rod2015}, the original name of the corresponding postulate is (CU1) and that of the semantic characterization is (CRU1). We rename them as (C1$\belsetUpd$) and (CR1$\belsetUpd$), respectively.}.

	\begin{description}
		\item[(C1$\belsetUpd$)]
		If $\varphi \models \phi$, then $\belSet \belsetUpd_{\phi} \varphi \equiv \belSet \belsetUpd \varphi$.
		
		\item[(CR1$\belsetUpd$)]
		If $w_1, w_2 \in [\phi]$, then $w_1 \leq^{\phi}_w w_2$ iff $w_1 \leq_w w_2$.
	\end{description}
	
	\looseness=-1
	The semantics of the update operator $\belsetUpd_{\phi}$ is the same as the semantics in Theorem \ref{thm:KM_semantic} except it uses the updated preorder $\leq^\phi_w$ instead of the initial one $\leq_w$.
	The update operator $\belsetUpd_{\phi}$ therefore satisfies postulates (U1)-(U8). % for one-shot belief update.

	\begin{theorem}[\cite{Rod2015}]
		Let $\belsetUpd_{\phi}$ be an update operator satisfying postulates (U1)-(U8).
		Then, the operator $\belsetUpd_{\phi}$ satisfies postulates (C1$\belsetUpd$) iff each of its corresponding faithful collective assignment $\leq^{\phi}_{w}$ satisfies (CR1$\belsetUpd$).
	\end{theorem}
	
	\looseness=-1
	In the case where $\varphi \models \phi$, postulate (C1$\belsetUpd$) requires that iteratively updating $\belSet$ by the information $\phi$ and its subsequent one $\varphi$ yields the same belief set as directly updating the $\belSet$ by $\varphi$.
	As for its corresponding semantics, for every possible world $w$, the preorder $\leq^{\phi}_{w}$ is required to coincide with $\leq_{w}$ on each pair of possible worlds in $[\phi]$.
	
	\looseness=-1
	We hereafter  illustrate the update process of postulate (C1$\belsetUpd$) with the following example.% the distinction between the postulate (C1$\belUpd$') and postulate (C1$\belUpd$). 
	
	\begin{example}[Postulate (C1$\belsetUpd$)] \label{exm:C1_Rod2015}
		Consider again Example \ref{exm:C1}.
		Let us provide an update operator satisfying postulate (C1$\belsetUpd$).
		The initial belief set $\belSet$ and two new information $\phi$ and $\varphi$ are the same as those in Example \ref{exm:C1}.
		That is, $\belSet \equiv \form{w_0} \equiv \neg b \land  \neg n \land  \neg t$, $\phi = b \lor n$ and $\varphi = n$. % and is associated with the belief set $\staTObel{\belStatus} = $.
		The partial preorder $\leq_{w_0}$ is
		\begin{itemize}
			\item $w_0 <_{w_0} w_2, w_4 <_{w_0} w_1, w_3, w_5, w_6, w_7$.
		\end{itemize}
		
		\looseness=-1
		By semantics (CR1$\belsetUpd$), the new preorder $\leq^{\phi}_{w_0}$ should coincide with $\leq_{w_0}$ on every pair of two possible worlds in $[\phi] \!=\! \set{w_2, w_3, w_4, w_5, w_6, w_7}$.
		Suppose that $\leq^{\phi}_{w_0}$ is defined as:
		
		\begin{itemize}
			\item $w_0 <^{\phi}_{w_0} w_1 <^{\phi}_{w_0} w_2, w_4 <^{\phi}_{w_0}  w_3, w_5, w_6, w_7$.
		\end{itemize}
		
		\looseness=-1
		As the belief set $\belSet$ involves only $w_0$, the preorders w.r.t. other possible worlds do not affect the update result. 
		Hence we do not adjust them.
		In addition, $[\varphi] = \set{w_2, w_3, w_6, w_7}$.
		It follows from (C1$\belsetUpd$) that $\belSet \belsetUpd_{\phi} \varphi \equiv \belSet \belsetUpd \varphi \equiv \form{w_2} \equiv \neg b \land  n  \land \neg t$. \qed
		
	\end{example}

	\looseness=-1
	As seen above, after receiving the new information $\phi$, the agent maintains the initial belief set $\belSet$, believing that nothing is on the left zone. 
	The order over possible worlds shifts from $\leq_{w_0}$ to $\leq^{\phi}_{w_0}$.
	As a result, the final updated belief $\belSet \belsetUpd_{\phi} \varphi$ is actually deduced from the initial belief set $\belSet$, the new order $\leq^{\phi}$ and the newly acquired information $\varphi$.
	It does not meet the basic requirement of iterated belief update, since it has no effect on the evolution of the belief set in light of the sequential new information.
	Clearly, it is more rational that the final belief should come from the latest belief set $\belSet \belsetUpd \phi$ rather than $\belSet$. %, the new order $\leq^{\phi}$ and the information $\varphi$.
	This is achieved through the development of postulate (C1$\belUpd$) in this paper.
	
	\looseness=-1
	We remark that the problems in \cite{Rod2015} occurs on not only the migration of postulate (C1$\belsetUpd$) but also all those of the remaining iterated postulates.
	Similarly, these deficiencies can be fixed via using the corresponding postulate proposed in this section instead.

	%\vspace*{-2mm}
	\section{(In)compatibility Results of \\ Iterated Belief Update} % Postulates in Belief Update}
	%\vspace*{-1mm}
	%\CHANGE{	
	\looseness=-1
	% In last section, we present several postulates for iterated update and offer their semantics characterizations.
	In this section, we are going to analyze the compatibility between the presented postulates for iterated update and the modified KM postulates. % (\ie, postulates (U1$\belUpd$)-(U5$\belUpd$) and (U8$\belUpd$)-(U9$\belUpd$)).
	
	\looseness=-1
	We hereafter provide a concrete update operator satisfying the modified KM postulates. % \QUESTION{(?? use the modified KM postulates)}
	Given a possible world $w$ and a sentence $\phi$ s.t. $w \in [\phi]$, we divide the entire set $\worldSet$ of possible worlds into three hierarchies: $\subWorldSetH^{w, \phi}_0 = \set{w}$, $\subWorldSetH^{w, \phi}_1 = [\phi] \setminus \set{w}$ and $\subWorldSetH^{w, \phi}_2 = [\neg \phi]$.
	
	\looseness=-1	
	\begin{definition}  \rm  \label{def:Iterated_operator}
		Let $\belStatus$ be a belief state associated with a belief set $\staTObel{\belStatus}$ and a collection $\leq^S$ of preorders over possible worlds.
		Let $\phi$ be a sentence.
		The operator $\belUpdOper_1$ is defined as: % follows: 
		\begin{enumerate}
			\item $[\staTObel{\belStatus \belUpdOper_1 \phi}] = \bigcup\limits_{w \in [{\staTObel{\belStatus}}]} \min([\phi], \leq^{\belStatus}_{w})$;
%			\item for every $w \in [\staTObel{\belStatus \belUpdOper_1 \phi}]$ and $w_1, w_2 \in \worldSet$, $w_1 \leq_{w}^{S \belUpdOper_1 \phi} w_2$ iff \DELETE{$w_1 \in \subWorldSetH^{w, \staTObel{\belStatus \belUpdOper_1 \phi}}_i$, $w_2 \in \subWorldSetH^{w, \staTObel{\belStatus \belUpdOper_1 \phi}}_j$} \CHANGE{$w_1 \in \subWorldSetH^{w, \phi}_i$, $w_2 \in \subWorldSetH^{w, \phi}_j$} and $i \leq j$. % where $\varphi = \staTObel{\belStatus \belUpdOper_1 \phi}$.
			\item for every $w \in [\staTObel{\belStatus \belUpdOper_1 \phi}]$ and $w_1, w_2 \in \worldSet$, $w_1 \leq_{w}^{S \belUpdOper_1 \phi} w_2$ iff $w_1 \in \subWorldSetH^{w, \phi}_i$, $w_2 \in \subWorldSetH^{w, \phi}_j$ and $i \leq j$. 
		\end{enumerate}
	\end{definition}

	\looseness=-1
	The above two conditions impose the constraints on the updated belief set $\staTObel{\belStatus \belUpdOper_1 \phi}$ and the subsequent update strategy $\leq^{\belStatus \belUpdOper_1 \phi}$, respectively.
	Clearly, condition (1) is the same as the semantics of the update operator.
	As a result, the operator $\belUpdOper_1$ satisfies postulates (U1$\belUpd$)-(U8$\belUpd$) and hence being a KM-compatible update operator.
	Condition (2) assigns to every possible world $w$ satisfying $\staTObel{\belStatus \belUpdOper_1 \phi}$ a partial preorder $\leq_{w}^{\belStatus \belUpdOper_1 \phi}$.
	To be specific, the preorder $\leq_{w}^{\belStatus \belUpdOper_1 \phi}$ exactly characterizes a binary relation over possible worlds $\worldSet$ divided into three hierarchies: $\subWorldSetH^{w, \phi}_0$, $\subWorldSetH^{w, \phi}_1$ and $\subWorldSetH^{w, \phi}_2$.
	That is, (1) the possible world $w$ is the most preferable, followed by the ones satisfying the new information $\phi$, and finally the remaining ones falsifying $\phi$;
	and (2) every two possible worlds in $[\phi] \setminus \set{w}$ (resp. $[\neg \phi]$) are equally plausible.

	\looseness=-1
	Recalling the scenario in Example \ref{exm:C4}, we illustrate the specific update process of the operator $\belUpdOper_1$ as follows.
	
	\begin{example} \label{exm:Iterated_operator}
		The initial belief state $\belStatus$ is associated with the belief set $[\staTObel{\belStatus}] = \set{w_0, w_4}$ and a collection $\leq^{\belStatus}$ of partial preorders as follows.
		
		\begin{itemize}
			\item $ w_0 <^{\belStatus}_{w_0} w_4  <^{\belStatus}_{w_0} w_1, w_2, w_3, w_5, w_6, w_7$.
			\item $ w_4 <^{\belStatus}_{w_4} w_0  <^{\belStatus}_{w_4} w_1, w_2, w_3, w_5, w_6, w_7$.
		\end{itemize}
		%		}
		The new information $\phi = b$ corresponds to the set of possible worlds $\set{w_4, w_5, w_6, w_7}$.
		Clearly, the minimal elements satisfying $\phi$ w.r.t. $\leq^{\belStatus}_{w_0}$ and $\leq^{\belStatus}_{w_4}$ are both the posssible world $w_4$.
		By the semantics of the update operator, we get that $\staTObel{\belStatus \belUpdOper_1 \phi} \equiv \form{w_4} \equiv b \land \neg n \land \neg m$.
		According to the definition of $\belUpdOper_1$, the updated belief state $\belStatus \belUpdOper_1 \phi$ would be assigned to a collection $\leq^{\belStatus \belUpdOper_1 \phi}$ of preorders.
		\begin{itemize}
			\item $w_4 <^{\belStatus \belUpdOper_1 \phi}_{w_4} w_5, w_6, w_7 <^{\belStatus \belUpdOper_1 \phi}_{w_4} w_0, w_1, w_2, w_3$
		\end{itemize}

		As seen above, the possible worlds of $\worldSet$ are splitted into three hierarchies:
		(1) the most preferable ones $\subWorldSetH^{w_4, \phi}_0 = \set{w_4}$ are exactly the current belief set, stating that only the book is on the left zone;
		(2) the second most preferable ones $\subWorldSetH^{w_4, \phi}_1 = \set{w_5, w_6, w_7}$ are the other three possible worlds satisfying $\phi$, stating that the book and at least one of the novel and the toy are on the left zone;
		(3) the least preferable ones $\subWorldSetH^{w_4, \phi}_2 = \set{w_0, w_1, w_2, w_3}$ are the possible worlds falsifying $\phi$, stating that the book is not on the left zone.
	
		Finally, the newly acquired information $\varphi = n$ corresponds to the set of possible worlds $\set{w_2, w_3, w_6, w_7}$.
		%It is easily verified that 
		The minimal elements satisfying $\varphi$ w.r.t. $\leq^{\belStatus \belUpdOper_1 \phi}_{w_4}$ are the two possible worlds $w_6$ and $w_7$.
		Hence, the final updated belief set $\staTObel{(\belStatus \belUpdOper_1 \phi)\belUpdOper_1 \varphi} = \form{w_6, w_7}$, which argues that both the book and novel are on the left zone.\qed
		
	\end{example}

	\looseness=-1
	The following theorem confirms the compatibility results between two iterated postulates and the KM postulates.
%	It is possible to exclude some inadequate update manner in the iteration by supplementing all of (C3$\belUpd$) and (C4$\belUpd$) into the KM framework.
	
%	We hereafter show that the update operator $\belUpdOper_1$ satisfies postulates (C3$\belUpd$) and (C4$\belUpd$). %, (Lex$\belUpd$) and (Ind$\belUpd$).
	
	\begin{theorem} \label{thm:compatibility_C3-4}
		The update operator $\belUpdOper_1$ satisfies postulates (C3$\belUpd$) and (C4$\belUpd$). %, (Lex$\belUpd$) and (Ind$\belUpd$).
	\end{theorem}	

%	\looseness=-1
%	Clearly, Theorem \ref{thm:compatibility_C3-4} confirms the compatibility results between those iterated postulates and the KM postulates.
%	It is possible to exclude some unreasonable update manner in the iteration by supplementing each of (C3$\belUpd$), (C4$\belUpd$), (Lex$\belUpd$) and (Ind$\belUpd$) into the KM framework.

	\looseness=-1
	% In belief revision, a compatibility result between the DP postulates and the AGM postulates was proved by \citeauthor{DarP1997}.
	In the seminal paper \cite{DarP1997}, it is proved that all of the DP postulates for iterated revision are compatible with the AGM paradigm (a basic framework for belief revision). 
	Unfortunately, we can not draw a similar conclusion in the context of belief update.
	The following theorem shows that neither postulate (C1$\belUpd$) nor (C2$\belUpd$) is compatible with the KM postulates.

	\begin{theorem} \label{thm:noOperator_C1-2}
		%Let $\belUpd$ be an update operator satisfying (U1$\belUpd$)-(U8$\belUpd$).
		There are a belief state $\belState$ associated with a collection $\leq^\belState$ of partial preorders and a sentence $\varphi$ s.t. no update operator $\belUpd$ satisfies (U1$\belUpd$)-(U8$\belUpd$) along with (C1$\belUpd$) (resp. (C2$\belUpd$)).
		%, in addition to any of postulates (C1$\belUpd$) and (C2$\belUpd$). % and (Nat$\belUpd$).			
		%Each of postulates (C1$\belUpd$), (C2$\belUpd$) and (Nat$\belUpd$) is inconsistent with the KM postulates.
		%		There exists a belief state $\belStatus$ and a sentence $\phi$ such that no update operator satisfies postulate each of (C1$\belUpd$), (C2$\belUpd$) and (Nat$\belUpd$).
	\end{theorem}
	
	%	\ADD{
	\looseness=-1
	The incompatibility is due to the following reason.
	On the one hand, each of postulates (C1$\belUpd$) and (C2$\belUpd$) requires that the two belief states $\belStatus \belUpd \varphi$ and $(\belStatus \belUpd \phi) \belUpd \varphi$ should associate with the same belief set given a specific new information $\varphi$.
	One the other hand, it is possible that the two sets $\leq^\belStatus$ and $\leq^{\belStatus \belUpd \phi}$ have different numbers of partial preorders, resulting in two different belief sets which $\belStatus \belUpd \varphi$ and $(\belStatus \belUpd \phi) \belUpd \varphi$ have.

\vspace*{-2mm}
\section{Discussions}
\vspace*{-1mm}
\looseness=-1 
Examples \ref{exm:C1}-\ref{exm:C4} justify the rationality of the DP postulates in iterated update scenario.
In the following, we provide two counterexamples to (C1$\belUpd$) and (C2$\belUpd$), respectively.

%We observed that some of the above counterexamples can be used as counterexamples to (C1$\belUpd$)-(C4$\belUpd$).

	\begin{table}[t] 
	\centering
	%\resizebox{\columnwidth}{!}{
		\footnotesize
		\begin{tabular}{ll|l|l|l|l|}
			\hline
			\multicolumn{2}{|l|}{possible worlds}                                                                      
			& $w_0$    & $w_1$    & $w_2$    & $w_3$  \\ \hline \hline % & $w_4$    & $w_5$    & $w_6$    & $w_7$   \\ \hline \hline
			\multicolumn{2}{|l|}{\textbf{c}lean}                                                                           
			& $\false$ & $\false$ & $\true$ & $\true$ \\ \hline %& $\true$  & $\true$  & $\true$  & $\true$ \\ \hline
			\multicolumn{2}{|l|}{\textbf{d}irty}                                                                            
			& $\false$ & $\true$ & $\false$  & $\true$ \\ \hline % & $\false$ & $\false$ & $\true$  & $\true$ \\ \hline
		%	\multicolumn{2}{|l|}{book\textbf{m}ark}                                                                                 
		%	& $\false$ & $\true$  & $\false$ & $\true$  %& $\false$ & $\true$  & $\false$ & $\true$ \\ \hline			
		\end{tabular}
%	}
	\vspace*{-3mm}
	\caption{
		%				A KM-compatible operator violating postulate (C3$\belUpd$) % in Example \ref{exm:C3}.
		The definition of possible worlds in Example \ref{exm:counterC1}.
	}
	\vspace*{-3mm}
	\label{tab:counterC1}
\end{table}

\begin{example}[Postulate (C1$\belUpd$)] \label{exm:counterC1}
	\looseness=-1
	We simplify Example \ref{exm:KM_irrational}, and now the table contains only one zone and some glasses that are clean or dirty are on the table.
	The proposition $c$ (resp. $d$) denotes that ``some clean (resp. dirty) glasses are on the table".
	%We denote by $c$ (resp. $d$) the proposition ``there are some clean (resp. dirty) glasses on the table".
	Each possible world is defined in Table \ref{tab:counterC1}.
	Initially, no glass is on the table.
	The initial belief set $\staTObel{\belStatus}$ is $\neg c \land \neg d \equiv \form{w_0}$.
	The collection $\leq^{\belStatus}$ of partial preorders associated to $\belStatus$ contains the following preorder:
	\begin{itemize}
		\item $w_0 <^{\belStatus}_{w_0} w_1 <^{\belStatus}_{w_0} w_2, w_3$.
	\end{itemize}
		
	\looseness=-1
	We first ask a robot to place some glasses on the table, that is, the new information $\phi = c \lor d$.
	According to the semantics of the update operator, we obtain the new belief set $\staTObel{\belStatus \belUpd \phi} \equiv \neg c \land d$, believing only dirty glasses are on the table.
	Afterwards, we require the robot to place some clean glasses on the table, that is, the new information $\varphi = c$.
	Similarly, the belief set $\staTObel{\belStatus \belUpd \varphi}$ is $c$. 
	%We consider the other scenario: some clean glasses are observed to be on the table after a change, that is, the new information $\phi = d$.
	Since $\varphi \models \phi$, postulate (DP1$\belUpd$) requires that $\staTObel{\belStatus \belUpd \phi \belUpd \varphi} \equiv \staTObel{\belStatus \belUpd \varphi} \equiv c$.
	However, there are dirty glasses on the table in the belief state $\belStatus \belUpd \phi$.
	The robot should not change the state of the glasses already on the table. % by affecting an effect $c$.
	It makes sense to believe $c \land d$ after receiving the new information $\varphi$ from the belief state $\belStatus \belUpd \phi$.
	%in $\staTObel{\belStatus \belUpd \varphi \belUpd \phi}$.

\end{example}

\begin{example} [Postulate (C2$\belUpd$)] \label{exm:counterC2}
	%We use the previous example to postulate (DP2$\belUpd$).
	\looseness=-1
	We simplify Example \ref{exm:KM_irrational}, and the only items on the table are a book and a novel.
	Initially, the book and the novel can be found in any zone of the table, that is, the initial belief set $\staTObel{\belStatus}$ is $\true$.
	We first instruct a robot to move the book and the novel to the left zone, which is described as the new information $\phi = b \land n$.
	The updated belief set $\staTObel{\belStatus \belUpd \phi}$ is therefore $b \land n$.
	The new instruction to place the novel on the right zone, which is the new information $\varphi = \neg n$, is issued to the robot.
	There is no reason to abandon the proposition $b$ stating that the book is on the left zone, however, imposed by (C2$\belUpd$), we have $\staTObel{(\belStatus \belUpd \phi) \belUpd \varphi} = \staTObel{\belStatus \belUpd \varphi} = \neg n$. \qed
\end{example}

\looseness=-1
Examples \ref{exm:counterC1} and \ref{exm:counterC2} show that (C1$\belUpd$) and (C2$\belUpd$) leads to counterintuitive results which they aim to avoid, respectively.
%is too restrictive to accommodate a range of rational update strategy and Example \ref{exm:counterC2} says that (C2$\belUpd$) leads to counterintuitive results.
Postulates (C1$\belUpd$) and (C2$\belUpd$), however, are able to rule out inadequate update behaviors as shown in Examples \ref{exm:C1} and \ref{exm:C2}, respectively.
We remark that although the DP postulates are controversial in iterated belief revision\footnote{The criticism comes from convincing counterexamples in which the DP postulates cause counterintuitive results. \citet{Meyer2001} and \citet{Sta2009} put forward counterexamples to postulate (C1). Several counterexamples to postulate (C2) are provided by \citet{Cantwell1999} and \citet{KonP2000}. In addition, postulate (C2) was discussed by \citet{Leh1995}, \citet{DelDL06} and \citet{JinT2007}. \citet{Han2016} offered two counterexamples to (C3) and (C4), respectively.}, they still are the cornerstone of iterated belief change.
%In addition, we remark that the DP postulates are controversial in iterated belief revision.
%Notwithstanding, the DP postulates still are a cornerstone of iterated belief revision.
Many subsequent work on iterated belief change are based on the DP postulates.
For example, belief contraction is a type of belief change that studies the problem of how to remove a certain information from a belief set. % $\belSet$  $\varphi$.
A revision operator can be obtained from a contraction operator via the Levi identity \cite{Levi1978} and vice versa via the Harper identity \cite{Harper1976}.
%A contraction operator can be 
\citet{ChopraGMW2008} proposed four postulates for iterated belief contraction via slightly modifying the DP postulates.
\citet{BoothC2019} considered the four postulates as the benchmark of iterated belief contraction and use them to evaluate their proposed iterated contraction operators.
Reasoning about actions, an important topic of knowledge representation and reasoning, studies the change of agents' beliefs due to the effect of action.
Belief revision were incorporated into the situation calculus, a well-known framework of reasoning about actions \cite{ShapiroPLL2011,FangL2013,SchweringLP2017}.
These works considered the DP postulates as key properties of iterated belief revision and verified the satisfaction of the DP postulates by their approaches so as to demonstrate the advantage of their approaches.
Hence, the four postulates (C1$\belUpd$)-(C4$\belUpd$) we propose in this paper is a cornerstone of the subsequent research to iterated belief update.

\vspace*{-2mm}
	%\ADD{
	\section{Conclusions}
	\vspace*{-1mm}
	\looseness=-1
	In this paper, we have investigated the iteration of belief update.
	% Inspired by \citeauthor{DarP1997}, we first modify the original KM postulates to redefine update function on belief states.
	Inspired by \citeauthor{DarP1997}, we first presented a modification of the KM postulates framework over belief states.
	With the help of belief states, we migrated the DP postulates for iterated revision to the belief update scenario, contributing to four iterated update postulates.
	Furthermore, we offered the exact semantic characterizations based on partial preorders for each of the resulting postulates.
	At last, the (in)compatibility results between the iterated update postulates and the KM postulates are provided.
	We showed that, unlike in revision, each of postulates (C1$\belUpd$) and (C2$\belUpd$) for iterated update is incompatible with the KM postulates.
	
	\looseness=-1
	Despite being the most influential approach to iterated belief revision, the DP postulates are still too liberal to rule out unintended revision operators. % that lead to undesired revision behavior.
	To strengthen the DP postulates, some additional postulates are proposed, for example, natural (Nat) postulate \cite{Bou1996}, lexicographic (Lex) postulate \cite{NayPP2003}, and independence (Ind) postulate \cite{BooM2006,JinT2007}. % and restrained (Res) postulate \cite{BooM2006}.
	We extended the above three postulates to iterated belief update.
	In addition, we studied their semantic characterization and proved the representation theorem.
	The (in)compatibility results between them and the KM postulates are also analyzed.	
	Due to space limitation, these results were presented in Appendix.

\bibliography{AAAI-2024}

\newpage
\appendix

\section{Extending Iterated Belief Update with Strengthened Iterated Postulates}
\looseness=-1

We first review the three strengthened iterated revision postulates: (Nat) postulate \cite{Bou1996}, (Lex) postulate \cite{NayPP2003}, and (Ind) postulate \cite{BooM2006,JinT2007}, and their corresponding semantic characterizations.

\begin{description}
	\item[(Nat)] If $\staTObel{\belState \staRe \phi} \models \neg \varphi$, then $\staTObel{(\belState \staRe \phi) \staRe \varphi} \equiv \staTObel{\belState \staRe \varphi}$.
	
	\item[(Lex)] If $\varphi \not \models \neg \phi$, then $\staTObel{(\belState \staRe \phi) \staRe \varphi} \models \phi$.
	
	\item[(Ind)] If $\staTObel{\belState \staRe \varphi} \not \models \neg \phi$, then $\staTObel{(\belState \staRe \phi) \staRe \varphi} \models \phi$.
	
%	\item[(Res)] If $\staTObel{\belState \staRe \phi} \models \neg \varphi$ and $\staTObel{\belState \staRe \varphi} \models \neg \phi$, then $\staTObel{(\belState \staRe \phi) \staRe \varphi} \models \neg \phi$.
\end{description}

The semantic counterparts of the three postulates and the representation theorem \cite{Bou1996,NayPP2003,BooM2006,JinT2007} are given in the following.
\begin{description}
	\item[(NatR)] If $w_1, w_2 \notin [\staTObel{\belState \staRe \phi}]$, then $w_1 \leq_{\belState} w_2$ iff $w_1 \leq_{\belState \staRe \phi} w_2$.
	\item[(LexR)] If $w_1 \in [\phi]$ and $w_2 \in [\neg \phi]$, then $w_1 <_{\belState \staRe \phi} w_2$.
	\item[(IndR)] If $w_1 \in [\phi]$ and $w_2 \in [\neg \phi]$, then $w_1 \leq_{\belState} w_2$ only if $w_1 <_{\belState \staRe \phi} w_2$.
\end{description}

\begin{theorem}
	Let $\staRe$ be a revision operator satisfying (R*1)-(R*6).
	Then, $\staRe$ satisfies (Nat) (resp. (Lex)/(Ind)) iff the operator and its corresponding faithful assignment satisfies (NatR) (resp. (LexR)/(IndR)).
\end{theorem}

\looseness=-1
We hereafter show their migration to iterated belief update.
Similarly, we denote the resulting iterated update postulates by (Nat$\belUpd$), (Lex$\belUpd$) and (Ind$\belUpd$), which can be obtained by replacing the revision operator $\belRe$ in the corresponding revision postulate by the update operator $\belUpd$.
The exact semantic characterizations based on partial preorders for these postulates are listed in the following.

\begin{description}
	\item[(NatR$\belUpd$)] For every $\subWorldSet \subseteq \worldSet$ s.t. $\min([\phi], \leq^\belStatus_w) \cap \subWorldSet = \emptyset$ holds for every $w \in [\staTObel{\belStatus}]$, the following hold
	\begin{description}
	\item[Forth] for every $w \in [\staTObel{\belStatus}]$ and $w'' \in \min(\subWorldSet,  \leq^\belStatus_w)$, there is $w' \in [\staTObel{\belStatus \belUpd \phi}]$ s.t. $w'' \in \min(\subWorldSet, \leq^{\belStatus \belUpd \phi}_{w'})$.
	\item[Back] \!for every $w' \!\in\! [\staTObel{\belStatus \belUpd \phi}]$ and $w'' \!\in\! \min(\subWorldSet\!,\! \leq^{\belStatus \belUpd \phi}_{w'})$, there is $w \in [\staTObel{\belStatus}]$ s.t. $w'' \in \min(\subWorldSet, \leq^{\belStatus}_{w})$.
	\end{description}
	
	\item[(LexR$\belUpd$)] For every $\subWorldSet \subseteq \worldSet$ s.t. $\subWorldSet \cap [\phi] \neq \emptyset$, we have $\min(\subWorldSet, \leq^{\belStatus \belUpd \phi}_{w'}) \subseteq [\phi]$ for every $w' \in [\staTObel{\belStatus \belUpd \phi}]$.
	
	\item[(IndR$\belUpd$)] For every $\subWorldSet \subseteq \worldSet$ s.t. $\min(\subWorldSet, \leq^\belStatus_w) \cap [\phi] \neq \emptyset$ holds for some $w \in [\staTObel{\belStatus}]$, we have $\min(\subWorldSet, \leq^{\belStatus \belUpd \phi}_{w'}) \subseteq [\phi]$ for every $w' \in [\staTObel{\belStatus \belUpd \phi}]$.

\end{description}

\looseness=-1
The semantics (NatR$\belUpd$) requires that for every subset $\subWorldSet$ of possible worlds such that every minimal element of $[\phi]$ w.r.t. $\leq^{\belStatus}$ is not in $\subWorldSet$, the minimal elements of $\subWorldSet$ w.r.t. $\leq^{\belStatus \belUpd \phi}$ ought to be the same as that w.r.t. the preorders $\leq^\belStatus$.
The semantics (LexR$\belUpd$) means that for every subset $\subWorldSet$ with at least one possible world of $\phi$, all of the minimal elements of $\subWorldSet$ w.r.t. the collection $\leq^{\belStatus \belUpd \phi}$ of preorders associated with the updated belief state should satisfy $\phi$.
The semantics (IndR$\belUpd$) restricts $\leq^{\belStatus \belUpd \phi}$ in a similar way as (LexR$\belUpd$), except that it focuses on a subset $\subWorldSet$ such that its minimal elements w.r.t. $\leq^{\belStatus}$ to contain at least one possible world satisfying $\phi$.
%While the semantics (ResR$\belUpd$) focuses on the case where $\subWorldSet$ satisfies the conditions: (1) $\subWorldSet$ contains nothing of the minimal elements of $[\phi]$ w.r.t. $\leq^{\belState}$; and (2) the minimal elements of $\subWorldSet$ w.r.t. $\leq^{\belState}$ contains no elements of $[\phi]$.
In such case, it is required that the minimal elements of $\subWorldSet$ w.r.t. the preorders $\leq^{\belState \belUpd \phi}$ contains no elements of $[\phi]$, neither.

\looseness=-1
Theorem \ref{thm:StrRepr} shows the correspondence between each of the above three postulates and its corresponding semantics, based on faithful collective assignments.

\begin{theorem} \label{thm:StrRepr}
	Let $\belUpd$ be an update operator satisfying (U1$\belUpd$)-(U8$\belUpd$).
	Then, $\belUpd$ satisfies (Nat$\belUpd$) (resp. (Lex$\belUpd$)/(Ind$\belUpd$)) iff the operator and its corresponding faithful collective assignment satisfies (NatR$\belUpd$) (resp. (LexR$\belUpd$)/(IndR$\belUpd$)).
\end{theorem}

\looseness=-1
We also investigate the affiliation among the DP postulates and the above postulates for iterated belief update.
Postulates (Lex$\belUpd$) and (Ind$\belUpd$) can be viewed as the reinforcements of (C3$\belUpd$) and (C4$\belUpd$).
Besides, postulate (Nat$\belUpd$) is stronger than (C2$\belUpd$).

\begin{theorem} \label{thm:Relation_IteratedPostulates}
	Let $\belUpd$ be an update operator satisfying (U1$\belUpd$)-(U8$\belUpd$). 
	Then, the following hold
	\begin{itemize}
		\item $\belUpd$ satisfies (Lex$\belUpd$) only if it satisfies (Ind$\belUpd$).
		\item $\belUpd$ satisfies (Ind$\belUpd$) only if it satisfies (C3$\belUpd$) and (C4$\belUpd$).
		\item $\belUpd$ satisfies (Nat$\belUpd$) only if it satisfies (C2$\belUpd$).
	\end{itemize} 
\end{theorem}

\looseness=-1
Finally, we provide some (in)compatibility results between the above iterated update postulates and the modified KM postulates.
%Consider again the update operator $\belUpdOper_1$ defined in Definition \ref{def:Iterated_operator}.
%We define a new update operator $\belUpdOper_2$ by a modification of $\belUpdOper_1$ where $\subWorldSetH^w_1 = [\neg \phi]$ and $\subWorldSetH^w_2 = [\phi] \setminus \set{w}$.

\begin{theorem} \label{thm:compatibility_Lex_Ind_Res}
	The update operator $\belUpdOper_1$ satisfies postulates (Lex$\belUpd$) and (Ind$\belUpd$).
\end{theorem}

\begin{theorem} \label{thm:noOperator_Nat}
	There are a belief state $\belState$ associated with a collection $\leq^\belState$ of partial preorders and a sentence $\varphi$ s.t. no update operator $\belUpd$ satisfies (U1$\belUpd$)-(U8$\belUpd$) along with (Nat$\belUpd$).
\end{theorem}

\looseness=-1
Theorem \ref{thm:compatibility_Lex_Ind_Res} confirms that postulates (Lex$\belUpd$) and (Ind$\belUpd$) are compatible with the modified KM postulates.
In contrast, Theorem \ref{thm:noOperator_Nat} shows the incompatibility between postulate (Nat$\belUpd$) and the modified KM postulates.

\section{Supplemental Proofs}

{\noindent \bf Proof of Theorem \ref{thm:ModifiedKM_semantic}}
\begin{proof}
%	(consistent belief state ??)
	
	($\Rightarrow$):
	Assume that $\belUpd$ satisfies postulates (U1$\belUpd$)-(U8$\belUpd$).
	For every belief state $\belStatus$, we construct a collection $\set{\leq^{\belStatus}_{w}}_{w \in [\staTObel{S}]}$ of binary relations over $\worldSet$ as follow.
	
	\begin{itemize}
		\item Base case ($\size{[\staTObel{\belStatus}]} = 1$): $w_1 \leq^{\belStatus}_{w} w_2$ iff $w_1 = w$, or $[\staTObel{\belStatus \belUpd \form{w_1, w_2}}] = \set{w_1}$.
		
		\item Inductive case ($\size{[\staTObel{\belStatus}]} > 1$): 

		By (U8$\belUpd$), $\staTObel{\belStatus \belUpd \varphi} \equiv \bigvee\limits_{w_i \in [\staTObel{\belStatus}]} \staTObel{\belStatus_i \belUpd \varphi}$ for some set of belief states $\set{S_1, \cdots, S_n}$ s.t. $\staTObel{\belStatus_i} \equiv \form{w_i}$.
		
		We let $\leq^{\belStatus}_{w} = \leq^{\belStatus_i}_{w_i}$ where $w = w_i$.			
	\end{itemize}
	
	We first show that each relation $\leq^{\belStatus}_{w}$ is a partial preorder faithful to $w$ for every belief state $S$ and every $w \in [\staTObel{S}]$.
	%			}
	When $\size{[\staTObel{\belStatus}]} > 1$, each $\leq^{\belStatus}_{w}$ is $\leq^{\belStatus'}_{w}$ for some $\belStatus'$ s.t. $\size{[\staTObel{\belStatus'}]} = 1$ and $w \in [\staTObel{\belStatus'}]$.
	Hence, it suffices to verify the base case where $\size{[\staTObel{\belStatus}]} = 1$ and $w \in [\staTObel{\belStatus}]$.
	
	\begin{itemize}
		
		\item Reflexivity:
		Clearly, $w \leq^{\belStatus}_{w} w$.
		Suppose that $w' \neq w$.
		By the postulates (U1$\belUpd$) and (U3$\belUpd$), we get that $\staTObel{\belStatus \belUpd \form{w'}} \equiv \form{w'}$.
		It follows from the definition of $\leq^{\belStatus}_{w}$ that $w' \leq^{\belStatus}_{w} w'$.
		
		\item Transitivity:			
		Assume that $w_1 \leq^{\belStatus}_{w} w_2$ and $w_2 \leq^{\belStatus}_{w} w_3$.
		We hereafter prove that $w_1 \leq^{\belStatus}_{w} w_3$.
		\begin{itemize}
			\item The case where $w_1 = w$: it directly follows from the definition of $\leq^{\belStatus}_{w}$.
			
			\item The case where $w_1 \neq w$ and $w_2 = w$: it is impossible since $\leq^{\belStatus}_{w}$ is faithful, which is independently verified later.
			
			\item The case where $w_1 \neq w$ and $w_2 \neq w$:
			Let $\varphi = \form{w_1, w_2, w_3}$.
%		\CHANGE{
			Since $w_1 \leq^{\belStatus}_{w} w_2$ and $w_1 \not = w$, it follows from the definition of $\leq^{\belStatus}_{w}$ that $[\staTObel{\belStatus \belUpd \form{w_1, w_2}}] = \set{w_1}$.
			By (U5$\belUpd$), $[\staTObel{\belStatus \belUpd \varphi} \land \form{w_1, w_2}] \subseteq [\staTObel{\belStatus \belUpd \form{w_1, w_2}}]$.
			So, $w_2 \notin [\staTObel{\belStatus \belUpd \varphi}]$.
			Similarly, as $w_2 \leq^{\belStatus}_{w} w_3$ and $w_2 \not = w$, we can obtain that $[\staTObel{\belStatus \belUpd \form{w_2, w_3}}] = \set{w_2}$ and $w_3 \notin [\staTObel{\belStatus \belUpd \varphi}]$.
			These, together with postulates (U1$\belUpd$) and (U3$\belUpd$), imply that $[\staTObel{\belStatus \belUpd \varphi}] = \set{w_1}$.
			Hence, $\staTObel{\belStatus \belUpd \varphi} \models \form{w_1, w_3}$.
			Besides, it follows from (U1$\belUpd$) that $\staTObel{\belStatus \belUpd \form{w_1, w_3}} \models \form{w_1, w_3}$ and $\form{w_1, w_3} \models \varphi$.
			By (U6$\belUpd$), we get that $[\staTObel{\belStatus \belUpd \form{w_1, w_3}}] = [\staTObel{\belStatus \belUpd \varphi}] = \set{w_1}$.
			it follows from the definition of $\leq^{\belStatus}_{w}$ that $w_1 \leq^{\belStatus}_{w} w_3$.
%		}
		\end{itemize}
		
		\item Faithfulness: Let $w'$ be a possible world distinct to $w$.
		Clearly, $w \leq^{\belStatus}_{w} w'$.
		% By (U2$\belUpd$) and (U3$\belUpd$), we get that $\staTObel{\belStatus \belUpd \form{w, w'}} = \Th(w)$.
		By (U2$\belUpd$), we get that $\staTObel{\belStatus \belUpd \form{w, w'}} \equiv \form{w}$.
		So $w' \not \leq^{\belStatus}_{w} w$.
	\end{itemize}
	
	Finally we prove that $[\staTObel{\belStatus \belUpd \varphi}] \!=\! \bigcup\limits_{w \in [\staTObel{\belStatus}]} \min([\varphi], \leq^{\belStatus}_{w})$.
	
	The case where $\varphi$ is inconsistent, $[\varphi] = \emptyset$: it directly follows from (U1$\belUpd$) that $[\staTObel{\belStatus \belUpd \varphi}] = \emptyset$.
	Therefore, $[\staTObel{\belStatus \belUpd \varphi}] = \bigcup\limits_{w \in [\staTObel{\belStatus}]} \min([\varphi], \leq^{\belStatus}_{w}) = \emptyset$.
	
	The case where $\varphi$ is consistent: we first show that $[\staTObel{\belStatus \belUpd \varphi}] = \min([\varphi], \leq^{\belStatus}_{w})$ when $\staTObel{\belStatus} \equiv \form{w}$.
	\begin{itemize}
		\item We prove that $[\staTObel{\belStatus \belUpd \varphi}] \subseteq \min([\varphi], \leq^{\belStatus}_{w})$ by contradiction.
		Assume that $w_1 \in [\staTObel{\belStatus \belUpd \varphi}]$ and $w_1 \not \in \min([\varphi], \leq^{\belStatus}_{w})$.
		There is $w_2 \in [\varphi]$ s.t. $w_2 <^{\belStatus}_{w} w_1$.
		By (U1$\belUpd$) and the definition of $\leq^{\belStatus}_{w}$, we get that $[\staTObel{\belStatus \belUpd \form{w_1, w_2}}] = \set{w_2}$.
		It follows from (U5$\belUpd$) that $[\staTObel{\belStatus \belUpd \varphi} \land \form{w_1, w_2}] \subseteq [\staTObel{\belStatus \belUpd (\form{w_1, w_2} \land \varphi)}] = [\staTObel{\belStatus \belUpd \form{w_1, w_2}}]$.
		It follows that $w_1 \notin [\staTObel{\belStatus \belUpd \varphi} \land \form{w_1, w_2}]$.
		Hence, $w_1 \notin [\staTObel{\belStatus \belUpd \varphi}]$, which contradicts with the assumption.
		
%	\CHANGE{
		\item We prove that $\min([\varphi], \leq^{\belStatus}_{w}) \subseteq [\staTObel{\belStatus \belUpd \varphi}]$.
		Assume that $w' \in \min([\varphi], \leq^{\belStatus}_{w})$, it remains to verify that $w' \in [\staTObel{\belStatus \belUpd \varphi}]$. 
		In the case where $w' = w$, we get that $\staTObel{\belStatus} \models \varphi$. %, since $[\staTObel{\belStatus}] = \set{w}$ and $w = w' \in [\varphi]$.
		It directly follows from (U2$\belUpd$) that $w' \in [\staTObel{\belStatus \belUpd \varphi}]$. %, since $[\staTObel{\belStatus \belUpd \varphi}] = [\staTObel{\belStatus}]  = \set{w}  = \set{w'}$.
		In the case where $w' \not = w$.
		Let $[\varphi] = \set{w_1, \dots, w_n}$.
		Clealy, $\varphi \equiv \form{w', w_1} \lor \cdots \lor \form{w', w_n}$.
		Since $w' \in \min([\varphi], \leq^{\belStatus}_{w})$, for every $w_i \in [\varphi]$, it holds that $w_i \not <^{\belStatus}_{w} w'$.
		Thus, we get that (1) $w' \leq^{\belStatus}_{w} w_i$; or (2) $w' \not \leq^{\belStatus}_{w} w_i$ and $w_i \not \leq^{\belStatus}_{w} w'$.
		For Case (1), it follows from the definition of $\leq^{\belStatus}_{w}$ that $w' \in [\staTObel{\belStatus \belUpd \form{w', w_i}}]$.
		For Case (2), it follows from the definition of $\leq^{\belStatus}_{w}$ that $[\staTObel{\belStatus \belUpd \form{w', w_i}}] \not = \set{w'} \not = \set{w_i}$.
		This, together with postulates (U1$\belUpd$) and (U3$\belUpd$), implies that $[\staTObel{\belStatus \belUpd \form{w', w_i}}] = \set{w', w_i}$, hence $w' \in [\staTObel{\belStatus \belUpd \form{w', w_i}}]$.
		Therefore, we can obtain that $w' \in [\staTObel{\belStatus \belUpd \form{w', w_1}} \land \cdots \land \staTObel{\belStatus \belUpd \form{w', w_n}}]$.
		From (U1$\belUpd$), it holds that $w' \in [\staTObel{\belStatus \belUpd (\form{w', w_1} \lor \cdots \lor \form{w', w_n})}] = [\staTObel{\belStatus \belUpd \varphi}]$.
%	}
		
	\end{itemize}

	We now consider the situation where $\size{[\staTObel{S}]} > 1$.
%\CHANGE{
	According to postulate (U8$\belUpd$) and the definition of $\leq^S_w$, there is a set of belief state $\set{S_1, \cdots, S_n}$ s.t. $[\staTObel{\belStatus \belUpd \varphi}] \!=\! 
	\bigcup\limits_{w_i \in [\staTObel{\belStatus}]} [\staTObel{\belStatus_i \belUpd \varphi}]  \!=\! 
	\bigcup\limits_{w_i \in [\staTObel{\belStatus}]} \min([\varphi], \leq^{\belStatus_i}_{w_i})  \!=\!
	\bigcup\limits_{w_i \in [\staTObel{\belStatus}]} \min([\varphi], \leq^{\belStatus}_{w_i}) $ and $\staTObel{\belStatus_i} \equiv \form{w_i}$.

	($\Leftarrow$):
	Assume that there is a faithful collective assignment that maps each belief state $\belStatus$ to a collection $\leq^{\belStatus}$ of partial preorders.
	We define an operator $\belUpd$ by $[\staTObel{\belStatus \belUpd \varphi}] = \bigcup\limits_{w \in [\staTObel{\belStatus}]} \min([\varphi], \leq^{\belStatus}_{w})$.
	It is easily verified that (U1$\belUpd$), (U3$\belUpd$) and (U4$\belUpd$) hold.
	In addition, when $\staTObel{\belStatus}$ is inconsistent, (U2$\belUpd$), (U5$\belUpd$), (U6$\belUpd$), (U7$\belUpd$) and (U8$\belUpd$) hold.
	In the following, we only prove that $\belUpd$ satisfies postulates (U2$\belUpd$), (U5$\belUpd$), (U6$\belUpd$), (U7$\belUpd$) and (U8$\belUpd$) when $\staTObel{\belStatus}$ is consistent.
	
	\begin{itemize}
		\item (U2$\belUpd$):
		Assume that $\staTObel{\belStatus} \models \varphi$.
	
		By postulate (U8$\belUpd$), which is independently proved later, $[\staTObel{\belStatus \belUpd \varphi}] = \bigcup\limits_{w_i \in [\staTObel{\belStatus}]} [\staTObel{\belStatus_i \belUpd \varphi}]$ for some set of belief states $\set{S_1, \cdots, S_n}$ s.t. $\staTObel{\belStatus_i} \equiv \form{w_i}$.

		It follows from the definition of $\belUpd$ that $[\staTObel{\belStatus_i \belUpd \varphi}] \!=\! \min([\varphi], \leq^{\belStatus_i}_{w_i})$.
		Since $\staTObel{\belStatus} \models \varphi$, $w_i \in [\varphi]$.
		By the faithfulness of $\leq^{\belStatus}_{w}$, we get that $[\staTObel{\belStatus_i \belUpd \varphi}] = \set{w_i}$.
		Therefore, $[\staTObel{\belStatus \belUpd \varphi}] = [\staTObel{\belStatus}]$.
		
		\item (U5$\belUpd$):
		In the case where $\staTObel{\belStatus \belUpd \varphi} \land \phi$ is inconsistent, (U5$\belUpd$) trivially holds.
		We now consider the converse case.
		Let $w_1 \in [\staTObel{\belStatus \belUpd \varphi} \land \phi]$.
		Then $w_1 \in [\staTObel{\belStatus \belUpd \varphi}]$.
		By the definition of $\belUpd$, there is $w \!\in\! [\staTObel{\belStatus}]$ s.t. $w_1 \!\in\! \min([\varphi], \leq^{\belStatus}_{w})$.
		Since $[\varphi \land \phi] \subseteq [\varphi]$ and $w_1 \in [\phi]$, we get that $w_1 \in \min([\varphi \land \phi], \leq^{\belStatus}_{w})$.
		It follows from the definition of $\belUpd$ that $w_1 \in [\staTObel{\belStatus \belUpd (\varphi \wedge \phi)}]$, hence $\staTObel{\belStatus \belUpd \varphi} \land \phi \models \staTObel{\belStatus \belUpd (\varphi \wedge \phi)}$.

		\item (U6$\belUpd$):
		Suppose that $\staTObel{\belStatus \belUpd \varphi} \models \phi$ and $\staTObel{\belStatus \belUpd \phi} \models \varphi$.
%		It remains to verify that $\staTObel{\belStatus \belUpd \varphi} \equiv \staTObel{\belStatus \belUpd \phi}$.
		We first prove that $[\staTObel{\belStatus \belUpd \varphi}] \subseteq [\staTObel{\belStatus \belUpd \phi}]$ by contradiction.		
		Assume that $w_1 \in [\staTObel{\belStatus \belUpd \varphi}]$ and $w_1 \notin [\staTObel{\belStatus \belUpd \phi}]$.
		Since $\staTObel{\belStatus \belUpd \varphi} \models \phi$, $w_1 \in [\phi]$.
		As $w_1 \notin [\staTObel{\belStatus \belUpd \phi}]$, by the definition of $\belUpd$, for every $w \in [\staTObel{\belStatus}]$, $w_1 \notin \min([\phi],\!\! \leq^{\belStatus}_{w})$.
		Thus, there exist $w_2 \in [\phi]$ s.t. $w_2 \in \min([\phi], \leq^{\belStatus}_{w})$ and $w_2 <^{\belStatus}_{w} w_1$.
		It follows that $w_2 \in [\staTObel{\belStatus \belUpd \phi}]$.
		Since $\staTObel{\belStatus \belUpd \phi} \models \varphi$, $w_2 \in [\varphi]$.
		As $w_2 <^{\belStatus}_{w} w_1$, we get that $w_1 \notin \min([\varphi], \leq^{\belStatus}_{w})$.
		It follows from the definition of $\belUpd$ that $w_1 \notin [\staTObel{\belStatus \belUpd \varphi}]$, which contradicts with that $w_1 \in [\staTObel{\belStatus \belUpd \varphi}]$.
		Similarly, it can be verified that $[\staTObel{\belStatus \belUpd \varphi}] \supseteq [\staTObel{\belStatus \belUpd \phi}]$.
	
		\item (U7$\belUpd$):
		Suppose that $\staTObel{\belStatus}$ is complete.
		Let $\staTObel{\belStatus} \equiv \form{w}$.
		We prove by contradiction.
		Assume that $w_1 \in [\staTObel{\belStatus \belUpd \varphi} \land \staTObel{\belStatus  \belUpd \phi}]$ and $w_1 \notin [\staTObel{\belStatus \belUpd (\varphi \lor \phi)}]$.
		By the definition of $\belUpd$, $w_1 \in \min([\varphi], \leq^\belStatus_w)$, $w_1 \in \min([\phi], \leq^\belStatus_w)$ and $w_1 \notin \min([\varphi \lor \phi], \leq^\belStatus_w)$.
		Clealy, there exist $w_2 \in [\varphi \lor \phi]$ s.t. $w_2 \in \min([\varphi \lor \phi], \leq^{\belStatus}_{w})$ and $w_2 <^{\belStatus}_{w} w_1$.
		In the case where $w_2 \in [\varphi]$, we get that $w_1 \notin \min([\varphi], \leq^\belStatus_w)$, which contradicts with that $w_1 \in \min([\varphi], \leq^\belStatus_w)$.
		In another case where $w_2 \in [\phi]$, we get that $w_1 \notin \min([\phi], \leq^\belStatus_w)$, which contradicts with that $w_1 \in \min([\phi], \leq^\belStatus_w)$.
%}
		
		\item (U8$\belUpd$):
		It follows from the definition of $\belUpd$ that $[\staTObel{\belStatus \belUpd \varphi}] = \bigcup\limits_{w_i \in [\staTObel{\belStatus}]} \min([\varphi], \leq^{\belStatus}_{w_i})$.
		From the faithfulness, for every $w_i \in [\staTObel{S}]$, there is a belief state $S_i$ s.t. $\staTObel{S_i} \equiv \form{w_i}$ and $\leq^{\belStatus}_{w_i} = \leq^{\belStatus_i}_{w_i}$.
		By the definition of $\belUpd$, $[\staTObel{\belStatus_i \belUpd \varphi}] = \min([\varphi], \leq^{\belStatus_i}_{w_i})$.
		Since $\leq^{\belStatus}_{w_i} = \leq^{\belStatus_i}_{w_i}$, we get that $[\staTObel{\belStatus_i \belUpd \varphi}] = \min([\varphi], \leq^{\belStatus}_{w_i})$.
		Therefore, $\staTObel{\belStatus \belUpd \varphi} \equiv \bigvee \limits_{w_i \in [\staTObel{\belStatus}]} \staTObel{\belStatus_i \belUpd \varphi}$ for some set of belief states $\set{S_1, \cdots, S_n}$ s.t. $\staTObel{\belStatus_i} \equiv \form{w_i}$.

	\end{itemize}
\end{proof}

{\noindent \bf Proof of Theorem \ref{thm:iterRepr}}
\begin{proof}
	\begin{itemize}
		\item {\bf (C1$\belUpd$) $\Leftrightarrow$ (CR1$\belUpd$).}	
		
		($\Rightarrow$):
		Assume that $\belUpd$ satisfies (C1$\belUpd$).
		We here only show that the forth part of (CR1$\belUpd$).
		The back part can be similarly shown.
		Let $\subWorldSet \subseteq [\phi]$, $w \in [\staTObel{\belStatus}]$ and $w'' \in \min(\subWorldSet, \leq^{\belStatus}_w)$.
		We define $\varphi$ as $\form{\subWorldSet}$.
		Clearly, $\varphi \models \phi$.
		By (C1$\belUpd$), $\staTObel{(\belStatus \belUpd \phi) \belUpd \varphi} \equiv \staTObel{\belStatus \belUpd \varphi}$.
		It directly follows from the semantics of the update operator that 
		\[\bigcup\limits_{w \in [{\staTObel{\belStatus}}]} \min([\varphi], \leq^{\belStatus}_{w}) = \bigcup\limits_{w' \in [{\staTObel{\belStatus \belUpd \phi}}]} \min([\varphi], \leq^{\belStatus \belUpd \phi}_{w'}).\]
		
		Clearly, $w'' \in \bigcup\limits_{w \in [{\staTObel{\belStatus}}]} \min([\varphi], \leq^{\belStatus}_{w})$.
		By the above equation, $w'' \in \bigcup\limits_{w' \in [{\staTObel{\belStatus \belUpd \phi}}]} \min([\varphi], \leq^{\belStatus \belUpd \phi}_{w'})$.
		Hence, there is $w' \in [\staTObel{\belStatus \belUpd \phi}]$ s.t. $w'' \in \min(\subWorldSet, \leq^{\belStatus \belUpd \phi}_{w'})$.
		
		($\Leftarrow$):
		Assume that $\leq^\belStatus$ and $\leq^{\belStatus \belUpd \phi}$ satisfy (CR1$\belUpd$).
		Suppose that $\varphi \models \phi$. 
		We here only show that $[\staTObel{\belStatus \belUpd \varphi}] \subseteq [\staTObel{(\belStatus \belUpd \phi) \belUpd \varphi}]$.
		The opposite direction $[\staTObel{(\belStatus \belUpd \phi) \belUpd \varphi}] \subseteq [\staTObel{\belStatus \belUpd \varphi}]$ can be similarly proved.
		%			Clearly, $[\varphi] \subseteq [\phi]$.
		Let $w'' \in [\staTObel{\belStatus \belUpd \varphi}]$.
		It follows from the semantics of the update operator that there is $w \in [\staTObel{\belStatus}]$ s.t. $w'' \in \min([\varphi], \leq^{\belStatus}_{w})$. 
		By the forth part of (CR1$\belUpd$), there is $w' \in [\staTObel{\belStatus \belUpd \phi}]$ s.t. $w'' \in \min([\varphi], \leq^{\belStatus \belUpd \phi}_{w'})$.
		So $w'' \in [\staTObel{(\belStatus \belUpd \phi) \belUpd \varphi}]$.
		
		\item {\bf (C2$\belUpd$) $\Leftrightarrow$ (CR2$\belUpd$).}				
		The proof is symmetric with the one above.
		
		\item {\bf (C3$\belUpd$) $\Leftrightarrow$ (CR3$\belUpd$).}
		
		($\Rightarrow$):
		Assume that $\belUpd$ satisfies (C3$\belUpd$).
		Let $\subWorldSet \subseteq \worldSet$ s.t. $\min(\subWorldSet, \leq^\belStatus_w) \subseteq [\phi]$ holds for every $w \in [\staTObel{\belStatus}]$.
		We define $\varphi$ as $\form{\subWorldSet}$.
		% By the semantics (F$\belUpd$), $\phi \in B(S \belUpd \varphi)$.
		% By (C3$\belUpd$), we obtain that $\phi \in B((S \belUpd \phi) \belUpd \varphi)$.
		By the semantics of the update operator, $\staTObel{\belStatus \belUpd \varphi} \models \phi$.
		By (C3$\belUpd$), we obtain that $\staTObel{(\belStatus \belUpd \phi) \belUpd \varphi} \models \phi$.
		By the semantics of the update operator again, for every $w' \in [\staTObel{\belStatus \belUpd \phi}]$, $\min(\subWorldSet,  \leq^{\belStatus \belUpd \phi}_{w'}) \subseteq [\phi]$.
		
		($\Leftarrow$):
		Assume that $\leq^\belStatus$ and $\leq^{\belStatus \belUpd \phi}$ satisfy (CR3$\belUpd$).
		Suppose that $\staTObel{\belStatus \belUpd \varphi} \models \phi$.
		By the semantics of the update operator, for every $w \in [\staTObel{\belStatus}]$, $\min([\varphi], \leq^\belStatus_w) \subseteq [\phi]$.
		By (CR3$\belUpd$), we obtain that for every $w' \in [\staTObel{\belStatus \belUpd \phi}]$, $\min([\varphi],  \leq^{\belStatus \belUpd \phi}_{w'}) \subseteq [\phi]$.
		Hence, $\staTObel{(\belStatus \belUpd \phi) \belUpd \varphi} \models \phi$.
		
		\item {\bf (C4$\belUpd$) $\Leftrightarrow$ (CR4$\belUpd$).}
		
		($\Rightarrow$):
		Assume that $\belUpd$ satisfies (C4$\belUpd$).
		Let $\subWorldSet \subseteq \worldSet$ s.t. $\min(\subWorldSet, \leq^\belStatus_w) \cap [\phi] \neq \emptyset$ for some $w \in [\staTObel{\belStatus}]$.
		We define $\varphi$ as $\form{\subWorldSet}$.
		By the semantics of the update operator, $\staTObel{\belStatus \belUpd \varphi} \not\models \neg \phi$.
		By (C4$\belUpd$), we obtain that $\staTObel{(\belStatus \belUpd \phi) \belUpd \varphi} \not\models \neg \phi$.
		By the semantics of the update operator again, there is $w' \in [\staTObel{\belStatus \belUpd \phi}]$, $\min(\subWorldSet,  \leq^{\belStatus \belUpd \phi}_{w'}) \cap [\phi] \neq \emptyset$.
		
		($\Leftarrow$):
		Assume that $\leq^\belStatus$ and $\leq^{\belStatus \belUpd \phi}$ satisfy (CR4$\belUpd$).
		Suppose that $\staTObel{\belStatus \belUpd \varphi} \models \neg \phi$.
		By the semantics of the update operator, there is $w \in [\staTObel{\belStatus}]$, $\min([\varphi] \leq^\belStatus_w) \cap [\phi] \neq \emptyset$.
		By (CR4$\belUpd$), we get that there is $w' \in [\staTObel{\belStatus \belUpd \phi}]$, $\min([\varphi], \leq^{\belStatus \belUpd \phi}_{w'}) \cap [\phi] \neq \emptyset$.
		Hence, $\staTObel{(\belStatus \belUpd \phi) \belUpd \varphi} \not\models \neg \phi$.
		
	\end{itemize}
\end{proof}

{\bf Proof of Theorem \ref{thm:compatibility_C3-4}}
\begin{proof}
	It directly follows from the fact that postulate (Lex$\belUpd$) entails postulates (C3$\belUpd$) and (C4$\belUpd$) and that $\belUpdOper_1$ satisfies postulate (Lex$\belUpd$), which are independently proved later in Theorem \ref{thm:Relation_IteratedPostulates} and Theorem \ref{thm:compatibility_Lex_Ind_Res} respectively,
%}
\end{proof}

{\bf Proof of Theorem \ref{thm:noOperator_C1-2}}
\begin{proof}
	We first consider postulate (C1$\belUpd$).
	We will construct a belief state $S$ associated with a collection $\leq^{\belStatus}$ of partial preorders and a sentence $\phi$ s.t. there does not exist any collection $\leq^{\belStatus \belUpd \phi}$ of partial preorders s.t. $\leq^{\belStatus}$ and $\leq^{\belStatus \belUpd \phi}$ satisfy (CR1$\belUpd$).
	
	Let $\varSet = \set{p_1, p_2, p_3}$ and $\worldSet = \set{w_1, \dots, w_8}$\footnote{We remark that this proof holds no matter what the truth assignment on each possible world $w_i$ is. Hence, we do not fix a truth assignment on each world.}.
	Let $\belStatus$ be a belief state with its associated belief set $[\staTObel{\belStatus}] = \set{w_1, w_2}$ and its assigned partial preorders $\leq^{\belStatus}_{w_1}$ and $\leq^{\belStatus}_{w_2}$ as follows.
	\begin{itemize}
		\item $\!\! w_1 \!<^{\belStatus}_{w_1}\!\! w_2 \!<^{\belStatus}_{w_1}\!\! w_3 \!<^{\belStatus}_{w_1}\!\! w_4 \!<^{\belStatus}_{w_1}\!\! w_6 \!<^{\belStatus}_{w_1}\!\! w_5 \!<^{\belStatus}_{w_1}\!\! w_7 \!<^{\belStatus}_{w_1}\!\! w_8$.
		\item $\!\! w_2 \!<^{\belStatus}_{w_2}\!\! w_1 \!<^{\belStatus}_{w_2}\!\! w_3 \!<^{\belStatus}_{w_2}\!\! w_5 \!<^{\belStatus}_{w_2}\!\! w_6 \!<^{\belStatus}_{w_2}\!\! w_4 \!<^{\belStatus}_{w_2}\!\! w_7 \!<^{\belStatus}_{w_2}\!\! w_8$.
	\end{itemize}

	Let $\phi = \form{w_3, w_4, w_5, w_6}$.		
	By the semantics of the update operator, $[\staTObel{\belStatus \belUpd \phi}] = \set{w_3}$.
	The collection $\leq^{\belStatus \belUpd \phi}$ contains only one partial preorder $\leq^{\belStatus \belUpd \phi}_{w_3}$.
	% Assume that $\belStatus$ controls $\belStatus \belUpd \phi$ on $\phi$.
	% We shall show that $\belStatus \belUpd \phi$ does not control $\belStatus$ on $\phi$.
	Assume that the forth part of (CR1$\belUpd$) holds.
	That is, for every $\subWorldSet \subseteq [\phi]$, $w \in [\staTObel{\belStatus}]$ and $w'' \in \min(\subWorldSet,  \leq^\belStatus_w)$, there is $w' \in [\staTObel{\belStatus \belUpd \phi}]$ s.t. $w'' \in \min(\subWorldSet, \leq^{\belStatus \belUpd \phi}_{w'})$.
	We will show that the back part of (CR1$\belUpd$) does not hold, that is, there is $\subWorldSet_3 \subseteq [\phi]$, $w' \in [\staTObel{\belStatus \belUpd \phi}]$ and $w'' \in \min(\subWorldSet_3,  \leq^{\belStatus \belUpd \phi}_{w'})$ s.t. $w'' \notin \min(\subWorldSet_3, \leq^{\belStatus}_{w})$ holds for every $w \in [\staTObel{\belStatus}]$.

	Let $\subWorldSet_0 = \set{w_4, w_5}$, $\subWorldSet_1 = \set{w_5, w_6}$ and $\subWorldSet_2 = \set{w_4, w_6}$.
	Since $\subWorldSet_0 \subseteq [\phi]$, $w_4 \in \min(\subWorldSet_0, \leq^{\belStatus}_{w_1})$, by the assumption above, we have $w_4 \in \min(\subWorldSet_0, \leq^{\belStatus \belUpd \phi}_{w_3})$.
	Similarly, as $w_5 \in \min(\subWorldSet_0, \leq^{\belStatus}_{w_2})$, we have $w_5 \in \min(\subWorldSet_0, \leq^{\belStatus \belUpd \phi}_{w_3})$.
	It holds that $w_5 \not <^{\belStatus \belUpd \phi}_{w_3} w_4$ and $w_4 \not <^{\belStatus \belUpd \phi}_{w_3} w_5$.
	Therefore, $w_4$ and $w_5$ are either equivalently plausible ($w_4 \approx^{\belStatus \belUpd \phi}_{w_3} w_5$) or incomparable ($w_4 \neq^{\belStatus \belUpd \phi}_{w_3} w_5$ and $w_5 \neq^{\belStatus \belUpd \phi}_{w_3} w_4$) w.r.t. $\leq^{\belStatus \belUpd \phi}_{w_3}$.	
%	either $w_4$ and $w_5$ or $w_4$ and $w_5$ are incomparable w.r.t. $\leq^{\belStatus \belUpd \phi}_{w_3}$.	
	Similarly, since $\subWorldSet_1 \subseteq [\phi]$, $w_5 \in \min(\subWorldSet_1, \leq^{\belStatus}_{w_2})$ and $w_6 \in \min(\subWorldSet_1, \leq^{\belStatus}_{w_1})$, we get that
	$w_5$ and $w_6$ are either equivalently plausible or incomparable w.r.t. $\leq^{\belStatus \belUpd \phi}_{w_3}$.	
%	 $w_6 \not <^{\belStatus \belUpd \phi}_{w_3} w_5$ and $w_5 \not <^{\belStatus \belUpd \phi}_{w_3} w_6$.
	In addition, as $\subWorldSet_2 \subseteq [\phi]$, $w_4 \in \min(\subWorldSet_2, \leq^{\belStatus}_{w_1})$ and $w_6 \in \min(\subWorldSet_2, \leq^{\belStatus}_{w_2})$, we get that
	$w_4$ and $w_6$ are either equivalently plausible or incomparable w.r.t. $\leq^{\belStatus \belUpd \phi}_{w_3}$. % $w_6 \not <^{\belStatus \belUpd \phi}_{w_3} w_4$ and $w_4 \not <^{\belStatus \belUpd \phi}_{w_3} w_6$.
%	Therefore, either $w_4 \approx^{\belStatus \belUpd \phi}_{w_3} w_5 \approx^{\belStatus \belUpd \phi}_{w_3} w_6$ or the three possible worlds $w_4$, $w_5$ and $w_6$ are incomparable w.r.t. $\leq^{\belStatus \belUpd \phi}_{w_3}$.
	
	Let $\subWorldSet_3 = \set{w_4, w_5, w_6}$.
	Clearly, $\subWorldSet_3 \subseteq [\phi]$.
	It can be verified that $\min(\subWorldSet_3, \leq^{\belStatus \belUpd \phi}_{w_3}) = \set{w_4, w_5, w_6}$, hence $w_6 \in \min(\subWorldSet_3, \leq^{\belStatus \belUpd \phi}_{w_3})$.
	By the back part of (CR1$\belUpd$), we get that $w_6 \in \min(\subWorldSet_3, \leq^{\belStatus}_{w_1})$ or $w_6 \in \min(\subWorldSet_3, \leq^{\belStatus}_{w_2})$.
	However, neither $w_6 \in \min(\subWorldSet_3, \leq^{\belStatus}_{w_1})$ nor $w_6 \in \min(\subWorldSet_3, \leq^{\belStatus}_{w_2})$, which is a contradiction.
%}	
	
	The proof for (C2$\belUpd$) is similar to the above case except that we take into consideration $\phi = \form{w_3}$. 
%	Since (Nat$\belUpd$) entails (C2$\belUpd$) (cf. Theorem \ref{thm:Relation_IteratedPostulates}), we can draw the same conclusion for (Nat$\belUpd$). 	
\end{proof}

{\bf Proof of Theorem \ref{thm:StrRepr}}
\begin{proof}
	\begin{itemize}
		\item {\bf (Nat$\belUpd$) $\Leftrightarrow$ (NatR$\belUpd$).}  
		
		($\Rightarrow$):
		Assume that $\belUpd$ satisfies (Nat$\belUpd$).
		We here only show that the forth part of (NatR$\belUpd$).
		The back part can be similarly shown.
		Let $\subWorldSet \subseteq \worldSet$ s.t. $\min([\phi], \leq^\belStatus_w) \cap \subWorldSet = \emptyset$ holds for every $w \in [\staTObel{\belStatus}]$.
		We define $\varphi$ as $\form{\subWorldSet}$.
		It follows from the semantics of the update operator that $\staTObel{\belStatus \belUpd \phi} \models \neg \varphi$.
		By (Nat$\belUpd$), $\staTObel{(\belStatus \belUpd \phi) \!\belUpd\! \varphi} \!\equiv \! \staTObel{\belStatus \belUpd \varphi}$.
		By the semantics of the update operator again, we get
		
		\[\bigcup\limits_{w \in [{\staTObel{\belStatus}}]} \min([\varphi], \leq^{\belStatus}_{w}) = \bigcup\limits_{w' \in [{\staTObel{\belStatus \belUpd \phi}}]} \min([\varphi], \leq^{\belStatus \belUpd \phi}_{w'}).\] 
		
		Hence, for every $w \in [\staTObel{\belStatus}]$ and $w'' \in \min(\subWorldSet,  \leq^\belStatus_w)$, there is $w' \in [\staTObel{\belStatus \belUpd \phi}]$ s.t. $w'' \in \min(\subWorldSet, \leq^{\belStatus \belUpd \phi}_{w'})$.
		
		($\Leftarrow$):
		Assume that $\leq^\belStatus$ and $\leq^{\belStatus \belUpd \phi}$ satisfy (NatR$\belUpd$).	
		Assume that $\staTObel{\belStatus \belUpd \phi} \models \neg \varphi$.
		We only show that $[\staTObel{\belStatus \belUpd \varphi}] \subseteq [\staTObel{(\belStatus \belUpd \phi) \belUpd \varphi}]$.
		The opposite direction $[\staTObel{(\belStatus \belUpd \phi) \belUpd \varphi}] \subseteq [\staTObel{\belStatus \belUpd \varphi}]$ can be similarly proved.
		Since $\staTObel{\belStatus \belUpd \phi} \models \neg \varphi$, it follows from the semantics of the update operator that for every $w \in [\staTObel{\belStatus}]$, $\min([\phi], \leq^{\belStatus}_{w}) \cap [\varphi] = \emptyset$.
		Let $w'' \in [\staTObel{\belStatus \belUpd \varphi}]$.
		From the semantics of the update operator again, there is $w \in [\staTObel{\belStatus}]$ s.t. $w'' \in \min([\varphi], \leq^{\belStatus}_{w})$. 
		By the forth part of (NatR$\belUpd$), there is $w' \in [\staTObel{\belStatus \belUpd \phi}]$ s.t. $w'' \in \min([\varphi], \leq^{\belStatus \belUpd \phi}_{w'})$.
		So $w'' \in [\staTObel{(\belStatus \belUpd \phi) \belUpd \varphi}]$.
		
		\item {\bf (Lex$\belUpd$) $\Leftrightarrow$ (LexR$\belUpd$).}	
		
		($\Rightarrow$):
		Assume that $\belUpd$ satisfies (Lex$\belUpd$).
		Let $\subWorldSet \subseteq \worldSet$ s.t. $\subWorldSet \cap [\phi] \neq \emptyset$.
		We define $\varphi$ as $\form{\subWorldSet}$.
		Clearly, $\varphi \not \models \neg \phi$.
		By (Lex$\belUpd$), $\staTObel{(\belStatus \belUpd \phi) \belUpd \varphi} \models \phi$.
		It directly follows from the semantics of the update operator that for every $w' \in [\staTObel{\belStatus \belUpd \phi}]$, $\min(\subWorldSet,  \leq^{\belStatus \belUpd \phi}_{w'}) \subseteq [\phi]$.
		
		($\Leftarrow$):
		Assume that $\leq^\belStatus$ and $\leq^{\belStatus \belUpd \phi}$ satisfy (LexR$\belUpd$).	
		% Let $\varphi \not \models \phi$.
		Let $\varphi \not \models \neg \phi$.
		Clearly, $[\varphi] \cap [\phi] \neq \emptyset$.
		By (LexR$\belUpd$), for every $w' \in [\staTObel{\belStatus \belUpd \phi}]$, $\min([\varphi],  \leq^{\belStatus \belUpd \phi}_{w'}) \subseteq [\phi]$.
		It follows from the semantics of the update operator that $\staTObel{(\belStatus \belUpd \phi) \belUpd \varphi} \models \phi$.

		\item {\bf (Ind$\belUpd$) $\Leftrightarrow$ (IndR$\belUpd$).}	
		
		($\Rightarrow$):
		Assume that $\belUpd$ satisfies (Ind$\belUpd$).
		Let $\subWorldSet \subseteq \worldSet$ s.t. $\min(\subWorldSet, \leq^\belStatus_w) \cap [\phi] \neq \emptyset$ for some $w \in [\staTObel{\belStatus}]$.
		We define $\varphi$ as $\form{\subWorldSet}$.
		It follows from the semantics of the update operator that $\staTObel{\belStatus \belUpd \varphi} \not \models \neg \phi$.
		By (Ind$\belUpd$), $\staTObel{(\belStatus \belUpd \phi) \belUpd \varphi} \models \phi$.
		By the semantics of the update operator again, for every $w' \in [\staTObel{\belStatus \belUpd \phi}]$, $\min(\subWorldSet,  \leq^{\belStatus \belUpd \phi}_{w'}) \subseteq [\phi]$.
		
		($\Leftarrow$):
		Assume that $\leq^\belStatus$ and $\leq^{\belStatus \belUpd \phi}$ satisfy (IndR$\belUpd$).	
		Assume that $\staTObel{\belStatus \belUpd \varphi} \not \models \neg \phi$.
		It follows from the semantics of the update operator that there is $w \in [\staTObel{\belStatus}]$, $\min([\varphi], \leq^\belStatus_w) \cap [\phi] \neq \emptyset$.
		By (IndR$\belUpd$), for every $w' \in [\staTObel{\belStatus \belUpd \phi}]$, $\min([\varphi],  \leq^{\belStatus \belUpd \phi}_{w'}) \subseteq [\phi]$.
		Hence, $\staTObel{(\belStatus \belUpd \phi) \belUpd \varphi} \models \phi$.

	\end{itemize}
\end{proof}

{\bf Proof of Theorem \ref{thm:Relation_IteratedPostulates}}
\begin{proof}
	\begin{itemize}
		\item {\bf (Lex$\belUpd$) $\Rightarrow$ (Ind$\belUpd$).}	
		By Theorem \ref{thm:StrRepr}, it suffices to verify that (LexR$\belUpd$) implies (IndR$\belUpd$).
		Let $\subWorldSet \subseteq \worldSet$ s.t. $\min(\subWorldSet, \leq^\belStatus_w) \cap [\phi] \neq \emptyset$ for some $w \in [\staTObel{\belStatus}]$.
		Clearly, $\subWorldSet \cap [\phi] \neq \emptyset$.
		By (LexR$\belUpd$), for every $w' \in [\staTObel{\belStatus \belUpd \phi}]$, $\min(\subWorldSet,  \leq^{\belStatus \belUpd \phi}_{w'}) \subseteq [\phi]$.
		
		\item {\bf (Ind$\belUpd$) $\Rightarrow$ (C3$\belUpd$).}	
		By Theorem \ref{thm:iterRepr} and Theorem \ref{thm:StrRepr}, it suffices to verify that (IndR$\belUpd$) implies (CR3$\belUpd$).
		Let $\subWorldSet \subseteq \worldSet$ s.t. $\min(\subWorldSet, \leq^\belStatus_w) \subseteq [\phi]$ holds for every $w \in [\staTObel{\belStatus}]$.
		Clearly, $\min(\subWorldSet, \leq^\belStatus_w) \cap [\phi] \neq \emptyset$.
		By (IndR$\belUpd$), for every $w' \in [\staTObel{\belStatus \belUpd \phi}]$, $\min(\subWorldSet,  \leq^{\belStatus \belUpd \phi}_{w'}) \subseteq [\phi]$.
		
		\item {\bf (Ind$\belUpd$) $\Rightarrow$ (C4$\belUpd$).}	
		By Theorem \ref{thm:iterRepr} and Theorem \ref{thm:StrRepr}, it suffices to verify that (IndR$\belUpd$) implies (CR4$\belUpd$).
		% Let $\subWorldSet \subseteq \worldSet$ and $w \in [\staTObel{\belStatus}]$ s.t. $\min(\subWorldSet, \leq^\belStatus_w) \cap [\phi] \neq \emptyset$.
		Let $\subWorldSet \subseteq \worldSet$ s.t. $\min(\subWorldSet, \leq^\belStatus_w) \cap [\phi] \neq \emptyset$ holds for some $w \in [\staTObel{\belStatus}]$.
		By (IndR$\belUpd$), for every $w' \in [\staTObel{\belStatus \belUpd \phi}]$, $\min(\subWorldSet,  \leq^{\belStatus \belUpd \phi}_{w'}) \subseteq [\phi]$.
		Therefore, $\min(\subWorldSet,  \leq^{\belStatus \belUpd \phi}_{w'}) \cap [\phi] \neq \emptyset$.
		
		\item {\bf (Nat$\belUpd$) $\Rightarrow$ (C2$\belUpd$).}	
		By Theorem \ref{thm:iterRepr} and Theorem \ref{thm:StrRepr}, it suffices to verify that (NatR$\belUpd$) implies (CR2$\belUpd$).
		Let $\subWorldSet \subseteq [\neg \phi]$.
		It follows that $\min([\phi], \leq^\belStatus_w) \cap \subWorldSet = \emptyset$ for every $w \in [\staTObel{\belStatus}]$.
		By (NatR$\belUpd$), we get that
		\[\bigcup\limits_{w \in [{\staTObel{\belStatus}}]} \min(\subWorldSet, \leq^{\belStatus}_{w}) = \bigcup\limits_{w' \in [{\staTObel{\belStatus \belUpd \phi}}]} \min(\subWorldSet, \leq^{\belStatus \belUpd \phi}_{w'}).\]
	\end{itemize}
\end{proof}

{\bf Proof of Theorem \ref{thm:compatibility_Lex_Ind_Res}}
\begin{proof}
	%We verify that $\belUpdOper_1$ satisfies postulate (Lex$\belUpd$).
	As a corollary of Theorem \ref{thm:Relation_IteratedPostulates}, postulate (Ind$\belUpd$) holds in $\belUpd_1$.
	By Theorem \ref{thm:StrRepr}, it suffices to verify that $\leq^{\belStatus}$ and  $\leq^{\belStatus \belUpd_1 \phi}$ satisfy (LexR$\belUpd$).
	Let $\subWorldSet \subseteq \worldSet$ s.t. $\subWorldSet \cap [\phi] \neq \emptyset$.
	Let $w' \in [\staTObel{\belStatus \belUpd_1 \phi}]$.
%	By the construction of the operator $\belUpdOper_1$ (Definition \ref{def:Iterated_operator}), for every $w_1 \in [\phi]$ and $w_2 \in [\neg \phi]$, $w_1 <^{\belStatus \belUpd_1 \phi}_{w'} w_2$.
	By the construction of the operator $\belUpdOper_1$ (cf. Definition \ref{def:Iterated_operator}), $w' \in [\phi]$ and for every $w_1 \in \subWorldSetH^{w', \phi}_0 \cup \subWorldSetH^{w', \phi}_1$ and $w_2 \in \subWorldSetH^{w', \phi}_2$, $w_1 <^{\belStatus \belUpd_1 \phi}_{w'} w_2$.
	That is, for every $w_1 \in [\phi]$ and $w_2 \in [\neg \phi]$, $w_1 <^{\belStatus \belUpd_1 \phi}_{w'} w_2$.
	This, together with $\subWorldSet \cap [\phi] \neq \emptyset$, implies that $\min(\subWorldSet, \leq^{\belStatus \belUpd_1 \phi}_{w'}) \subseteq [\phi]$.
\end{proof}

{\bf Proof of Theorem \ref{thm:noOperator_Nat}}
\begin{proof}
	It directly follows from the fact that (Nat$\belUpd$) entails (C2$\belUpd$) (cf. Theorem \ref{thm:Relation_IteratedPostulates}) and Theorem \ref{thm:noOperator_C1-2}. 
\end{proof}

\end{document}